%% file: main.tex
\documentclass[preprint,12pt]{elsarticle}

\usepackage{amssymb}
\usepackage{amsmath}

\usepackage{booktabs}
\usepackage{multirow}
\usepackage{xcolor}
\usepackage{hyperref}
\usepackage[all]{nowidow}
\usepackage{microtype}

\begin{document}

\begin{frontmatter}


\title{Convolutional Set Transformer}

\author[inst1]{Federico Chinello\corref{cor1}}
\ead{federico.chinello@studbocconi.it}

\author[inst2]{Giacomo Boracchi}
\ead{giacomo.boracchi@polimi.it}

\cortext[cor1]{Corresponding author.}

\affiliation[inst1]{
department={Dep. of Computing Sciences,},
organization={Bocconi University},
country={Italy}
}

\affiliation[inst2]{
department={Dep. of Electronics, Information and Bioengineering,}, 
organization={Politecnico di Milano},
country={Italy}
}

\begin{abstract}
We introduce the Convolutional Set Transformer (CST), a novel neural architecture designed to process image sets of arbitrary cardinality that are visually heterogeneous yet share high-level semantics—such as a common category, scene, or concept. Existing set-input networks, e.g., Deep Sets and Set Transformer, are limited to vector inputs and cannot directly handle 3D image tensors. As a result, they must be cascaded with a feature extractor, typically a CNN, which encodes images into embeddings before the set-input network can model inter-image relationships. In contrast, CST operates directly on 3D image tensors, performing feature extraction and contextual modeling \textit{simultaneously}, thereby enabling synergies between the two processes. This design yields superior performance in tasks such as Set Classification and Set Anomaly Detection and further provides native compatibility with CNN explainability methods such as Grad-CAM, unlike competing approaches that remain opaque. Finally, we show that CSTs can be pre-trained on large-scale datasets and subsequently adapted to new domains and tasks through standard Transfer Learning schemes.  To support further research, we release CST-15, a CST backbone pre-trained on ImageNet (\href{https://github.com/chinefed/convolutional-set-transformer}{link}).
\end{abstract}




\end{frontmatter}



\input{sec/1_introduction}
\input{sec/2_problem_formulation}

\input{sec/3_related_work}
\input{sec/4_proposed_solution}
\input{sec/5_experiments}

\input{sec/6_conclusion}

\bibliography{main}
\bibliographystyle{elsarticle-num.bst}
\newpage

\appendix
\input{appendix.tex}

\end{document}

%% file: sec/1_introduction.tex
\section{Introduction}
\label{sec:introduction}

\begin{figure*}[ht]
    \centering
\includegraphics[width=0.95\textwidth]{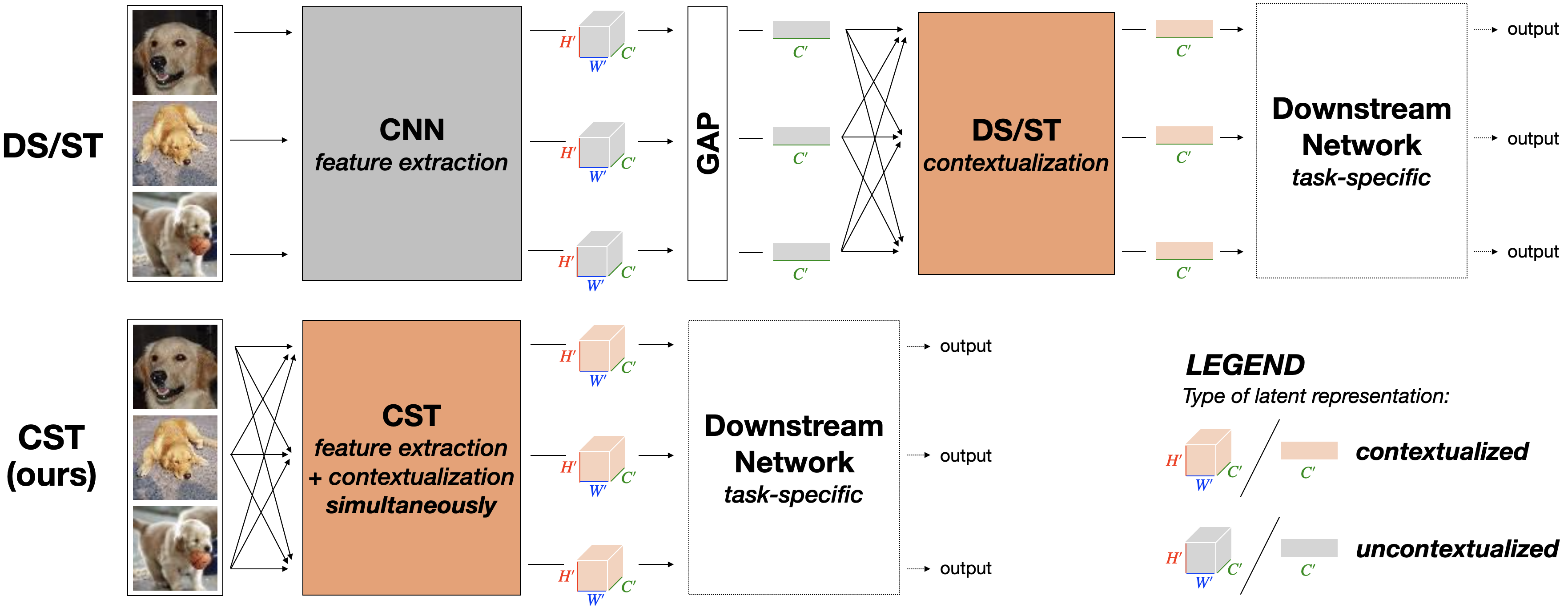}
    \caption{\textbf{Deep Sets (DS) and Set Transformer (ST) pipelines compared to the CST architecture (ours).} In DS/ST pipelines, a shared CNN is applied independently to each image in the input set, yielding a set of \textit{uncontextualized activation volumes} whose spatial dimensions are subsequently collapsed via Global Average Pooling, resulting in a set of \textit{uncontextualized latent vectors}. These vectors are then fed to the DS/ST module, which performs contextualization and outputs a set of \textit{contextualized latent vectors} for downstream processing. In contrast, CST (ours) processes the set of images through a stack of SetConv2D blocks, simultaneously performing feature extraction and context modeling, and yielding \textit{contextualized activation volumes} that retain spatial dimensions $(H' \times W' \times C')$ and can be directly used in downstream tasks.}
    \label{fig:comparison}
\end{figure*}

Images are ubiquitous and often occur in sets. For example, images on a web page or in user-generated social media content are typically related to a common concept and can therefore be processed together to enhance automatic visual recognition. In medicine, jointly analyzing multiple images from the same patient can improve diagnostic accuracy, such as in melanoma detection \cite{collenne2024reset}. Similarly, \cite{black2024multi} demonstrates how sets of room images can be used for fine-grained hotel classification, a task that supports efforts to combat human trafficking. In disaster response, collections of images depicting the same building from different viewpoints provide valuable input for automated damage assessment models \cite{khajwal2023post}. Overall, processing images as sets is advantageous because it enables visual recognition tasks to exploit both the complementary information carried by individual images and the semantic relationships that link them.

\begin{figure*}[!ht]
    \centering
    \includegraphics[width=0.8\textwidth]{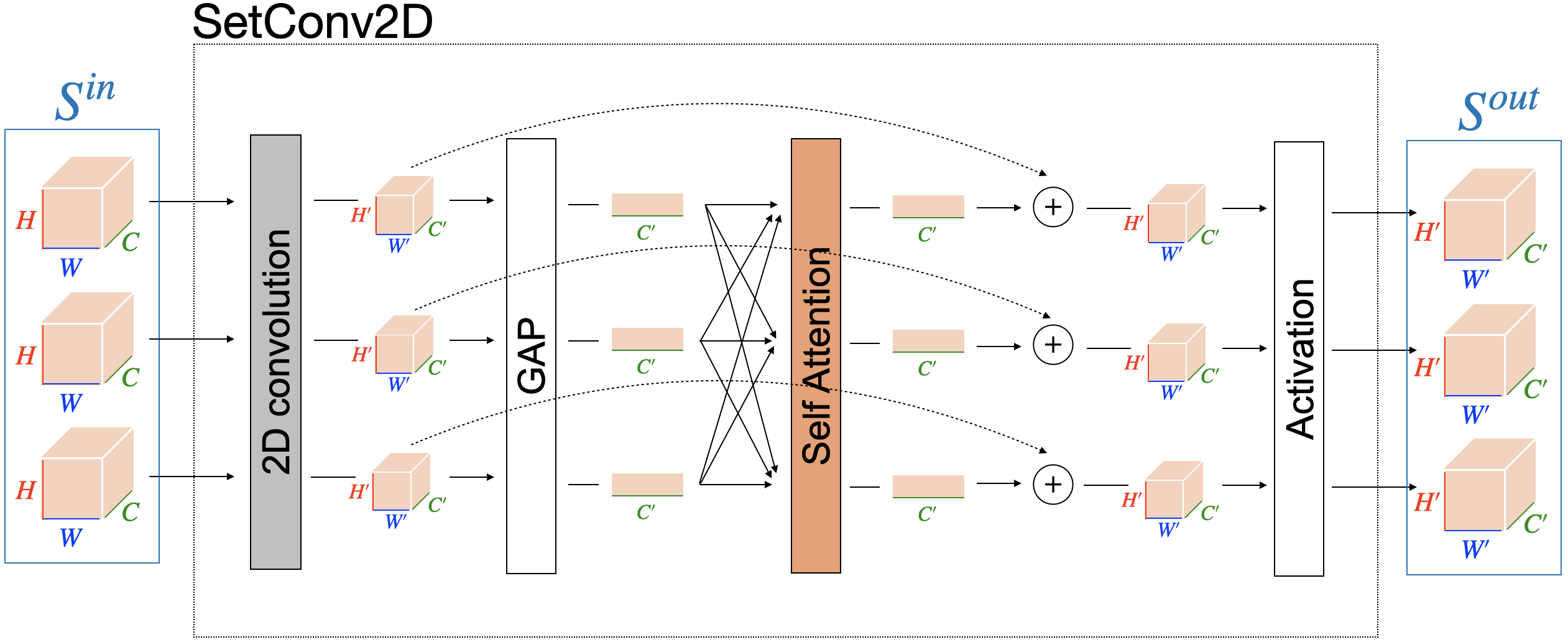}
    \caption{\textbf{SetConv2D layer}. SetConv2D takes a set of volumes as input and processes each of these by a shared convolutional layer. A Global Average Pooling (GAP) is applied to each volume to yield a set of latent vectors that is fed to a Multi-Head Self-Attention (MHSA) unit. The output of MHSA is used as context-aware bias vectors that are summed to the volumes fed to the GAP, via a residual connection. After applying these biases, the volumes are passed through a non-linear activation function.}
    \label{fig:setconv2d}
\end{figure*}

Unlike matrices or tensors, sets have no fixed cardinality and lack an inherent ordering of their elements. Convolutional Neural Networks (CNNs) have achieved remarkable success across a wide range of computer vision tasks, but they cannot natively process sets or exploit the contextual relationships among images within a set. By contrast, set-input neural networks such as Deep Sets \cite{zaheer2017deep} and Set Transformer \cite{lee2019set} are explicitly designed to handle set-structured data. However, these models operate only on sets of vectors and cannot directly process sets of 3D tensors such as images. As a result, each image in a set must first be converted into an embedding vector by a feature extractor—typically a CNN—and only then passed to the set-input network, as illustrated in Figure \ref{fig:comparison}. In this cascaded pipeline, the feature extractor processes each image independently and identically, without leveraging any contextual information, while the set-input network subsequently applies set-specific operations to capture inter-image relationships. While end-to-end training of the CNN and the Deep Sets or Set Transformer is possible, feature extraction and context modeling remain strictly sequential. As a result, the two processes cannot develop true synergies: convolutional layers enforce locality and translation-invariance priors during feature extraction, but these inductive biases cannot interact directly with the set-specific operations that capture semantic relationships among images, since the latter only occur once feature extraction has been completed. This cascaded design also limits explainability: saliency methods such as Grad-CAM \cite{Selvaraju_2017_ICCV} can only be applied to the convolutional layers, which precede any set-specific operation.

\begin{figure*}[ht]
    \centering
    \includegraphics[width=0.90\textwidth]{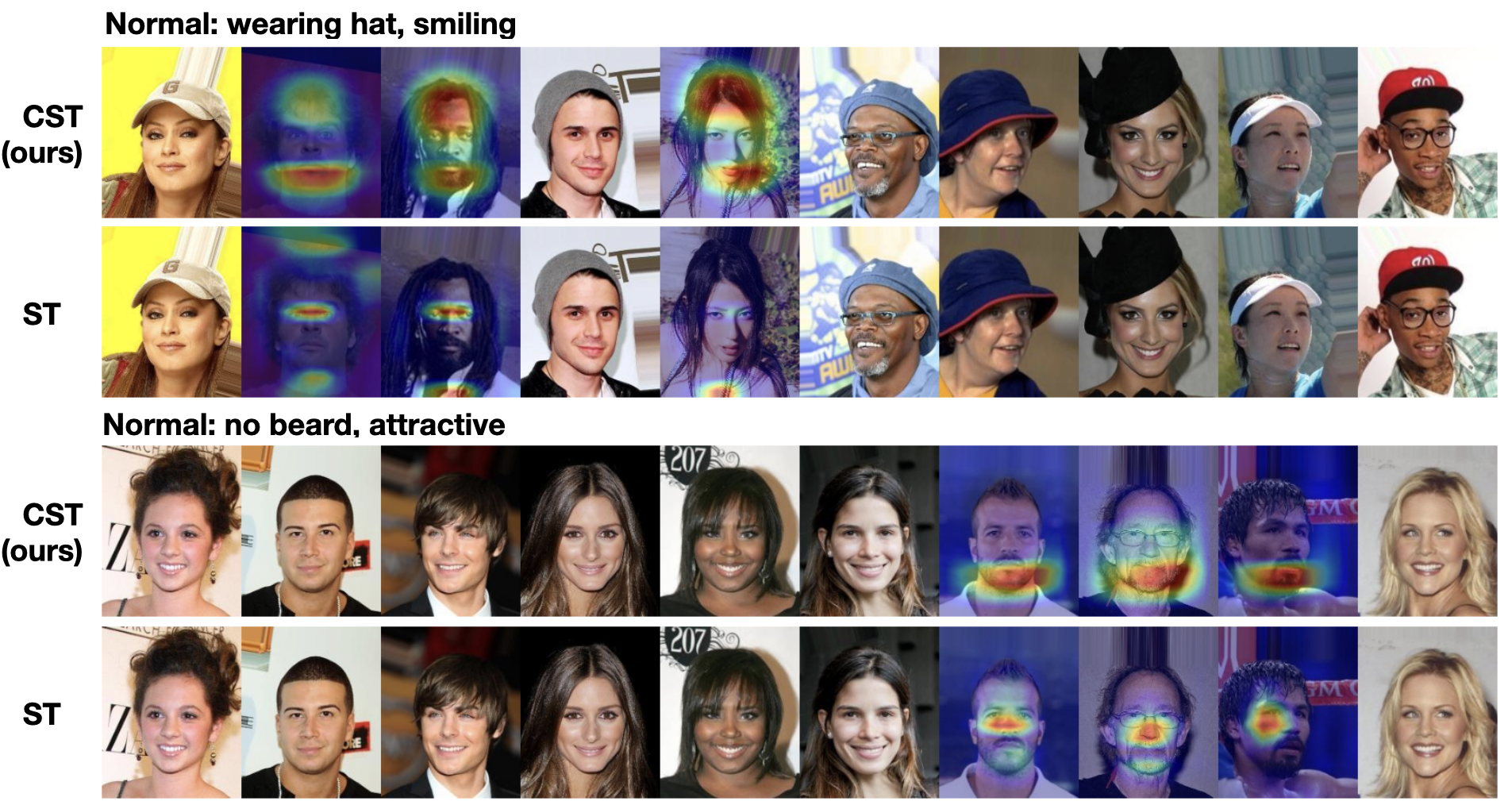}
    \caption{\textbf{Unlike Set Transformer (ST), CST is transparent and explainable}. The Set Anomaly Detection task aims to identify anomalous images within a set.  The figure shows two image sets derived from the CelebA dataset \cite{liu2015faceattributes}. In each set, a majority of normal images share two attributes (\textit{wearing hat} and \textit{smiling} in the first, \textit{no beard} and \textit{attractive} in the second), while a minority lack these attributes and are thus anomalous. After training a CST and an ST on CelebA for  Set Anomaly Detection, we evaluate the explainability of their predictions by overlaying Grad-CAMs on anomalous images. CST explanations correctly highlight the anomalous regions, whereas ST explanations fail to provide meaningful insights. For further details, see Section \ref{sec:anomaly}.}
    \label{fig:gradcam_anomaly}
\end{figure*}

To address these limitations, we introduce the Convolutional Set Transformer (CST), a general framework for deep learning architectures specifically designed for sets of images. Its fundamental building block is SetConv2D (see Figure \ref{fig:setconv2d}), a permutation-equivariant set-to-set layer that takes as input a set of 3D tensors—such as images or activation volumes—and returns a corresponding set of activation volumes, each primarily associated with one input image. The key distinction between Conv2D and SetConv2D is that, in the latter, biases are dynamically adjusted at inference time to incorporate contextual information extracted from the entire image set. These dynamic biases are computed by a Multi-Head Self-Attention (MHSA) \cite{vaswani2017attention} unit integrated into each SetConv2D layer. CST networks are obtained by stacking multiple SetConv2D layers and, unlike Deep Sets and Set Transformers, can exploit contextual information from the set during feature extraction (Figure \ref{fig:comparison}). In particular, dynamic biases in the early SetConv2D layers model relationships between images by analyzing low-level patterns, while in deeper layers they capture increasingly complex semantic connections.

Our experiments show that CST consistently outperforms Deep Sets and Set Transformer across multiple tasks, including Set Classification and Set Anomaly Detection. Moreover, unlike Deep Sets and Set Transformer, CST is natively compatible with explainability methods for CNNs, such as Grad-CAM \cite{Selvaraju_2017_ICCV}. In particular, we demonstrate that CST enables meaningful Grad-CAM visualizations in Set Anomaly Detection, while Set Transformer fails to provide reliable explanations (Figure \ref{fig:gradcam_anomaly}). Finally, whereas prior set learning literature, such as \cite{zaheer2017deep, lee2019set}, has mainly relied on training models from scratch on small, low-resolution datasets, we train a CST on a large-scale benchmark (ImageNet \cite{russakovsky2015imagenet}) and demonstrate that, once pre-trained, it can be effectively adapted to new domains via Transfer Learning. As a case study, we address the challenging task of event recognition from personal photo albums \cite{bossard13}.

We summarize the key contributions of our paper:
    \begin{itemize}
    \item \textbf{We introduce the Convolutional Set Transformer (CST)}, a deep neural architecture designed to operate on sets of images, and the SetConv2D block, the foundational module of CST.
    \item \textbf{We demonstrate that CST can be pre-trained on large-scale datasets} such as ImageNet and then adapted to a variety of downstream tasks through Transfer Learning. To this end, we introduce \textit{Contextualized Image Classification} (CIC), a natural and effective pre-training objective for CST, and \textit{Combinatorial Training} (CT), a training strategy designed to be used in conjunction with CIC. As a concrete contribution, we publicly release CST-15, a CST encoder pre-trained on ImageNet, for download.
    \item \textbf{We show that CST is natively compatible with standard CNN explainability tools} such as Grad-CAM, while existing set-input architectures—such as Deep Sets and Set Transformer—are inherently opaque due to their cascaded design.
\end{itemize}

%% file: sec/2_problem_formulation.tex
\section{Problem Formulation}
\label{sec:problem-formulation}
We now present the specific problems addressed in this paper. We address two broad categories of visual recognition tasks that operate on an input set of images \mbox{$\{I_1, \ldots, I_N \mid I_i \in \mathbb{R}^{H \times W \times C}\}$}: \textit{set-to-set} tasks and \mbox{\textit{set-to-global} tasks}.

\begin{itemize}
    \item In \textbf{set-to-set tasks}, the goal is to produce an output for each individual image while exploiting the contextual relationships among the images in $\{I_1, \ldots, I_N\}$. Within this category, we consider two tasks: \textit{Set Anomaly Detection} and \textit{Contextualized Image Classification} (CIC). Set Anomaly Detection is a binary classification task meant to identify images in a set that are anomalous or inconsistent with the majority of the set. Here, the notion of anomaly is relative: the same image may be considered anomalous in one set but not in another, depending on the surrounding context. In Contextualized Image Classification---a task we introduce in this paper---we assume that all \mbox{images in $\{I_1, \ldots, I_N\}$} share a common (unknown) label \(y \in \Lambda\). Rather than inferring this label independently for each image, we generate \textit{contextualized image-level predictions}, where the prediction for any given image is influenced by the other images in the same set. We propose CIC as an effective \mbox{pre-training task for CSTs}.
    \item In contrast, \textbf{set-to-global} tasks aim to generate a shared, global output for the entire set, such as in \textit{Set-level Classification}, where a single label $y \in \Lambda$ is predicted for the entire set \mbox{$\{I_1, \ldots, I_N\}$}.
\end{itemize}

For Set Classification tasks, the training dataset contains sets of images, where each set has an associated class label. For Set Anomaly Detection, following \cite{zaheer2017deep, lee2019set}, the training dataset consists of image sets in which images that are anomalous with respect to the surrounding context are marked. At both training and inference time, sets may contain any number of images, and their order is irrelevant. Classification tasks can also be performed on \textit{singleton sets} containing only a single image.

Set-input networks are either \textit{equivariant} or \textit{invariant} with respect to set permutations \cite{zaheer2017deep, lee2019set}. This distinction aligns with nature of the tasks discussed: permutation equivariance is essential for set-to-set tasks, where a permutation of the input set must result in a corresponding permutation of the output set. In contrast, permutation invariance is required for set-to-global tasks, ensuring that the output remains unaffected by any permutation of \mbox{the input set}.

\label{sec:image_set_classification}

%% file: sec/3_related_work.tex
\section{Related Work}

\label{sec:related_work}
Here, we present the literature relevant to our contributions, specifically set-input neural networks and networks that process multiple images as input.

\textbf{Learning from sets}. In the last decade, set-input deep neural networks have attracted significant interest from the research community \cite{zaheer2017deep, lee2019set, chenstacking, wagstaff2019limitations, bueno2021representation, wagstaff2022universal, naderializadeh2021pooling, bartunov2022equilibrium, zhang2019fspool} and found application to a broad range of tasks, including inference from point clouds \cite{qi2017pointnet, qi2017pointnet++}. 

Deep Sets \cite{zaheer2017deep} and Set Transformer \cite{lee2019set} are deep learning architectures designed to operate on set-structured data. However, they are restricted to processing sets of vectors and cannot manage sets of 3D tensors. Therefore, if the input consists of image sets, images must be encoded as embedding vectors by a conventional feature extractor, such as a CNN, before they can be fed into the Deep Sets/Set Transformer network (Figure \ref{fig:comparison}). In contrast, CST models contextual information \textit{simultaneously with feature extraction}, preserving spatial information, and leveraging the locality and translation invariance priors of images. Our experiments demonstrate that CST achieves superior performance and enables using explainability tools for CNNs on each individual image of the set. In contrast, Deep Sets and Set Transformer networks are opaque and their predictions are not always explainable.

Deep Sets for Symmetric elements (DSS) \cite{maron2020learning} is another architecture designed to operate on sets, but it is tailored for a fundamentally different class of problems than CST. In the case of image set processing, a DSS layer applies a shared convolutional layer $L_1$ to each image in the input set, while simultaneously computing the spatial summation of all images and processing it through a second convolutional layer $L_2$. The final output is then obtained by adding together the contributions from $L_1$ and $L_2$. Importantly, the use of spatial summation assumes a high degree of spatial alignment among the set elements, such as when the set contains shifted versions of the same image, noisy augmentations, or multi-channel variants of a common signal. This strong inductive bias limits DSS to tasks where such spatial alignment holds, like image restoration and denoising. DSS cannot capture the abstract semantic relationships that arise when a set contains visually diverse but conceptually related images. In contrast, CST is explicitly designed for these heterogeneous settings, where summing aligned features would be either ineffective or inappropriate. By combining Global Average Pooling with Multi-Head Self-Attention \cite{vaswani2017attention} within the SetConv2D framework, CST is able to exploit high-level semantic relationships across diverse set elements. As a result, CST can address complex tasks such event recognition from personal photo albums—applications that directly violate the core spatial alignment prior assumed by DSS.

\textbf{Learning from multiple images}. Multi-image deep networks have demonstrated their effectiveness in several domains, including 3D shape recognition \cite{su2015multi, wei2020view, liang2021mhfp, liu2022vfmvac, yang2025webly}, biomedical image analysis \cite{lin2025multi, sun2019multi, liu2018multi, van2021multi, black2024multi}, anomaly detection \cite{zhang2024attention}, and plant species identification \cite{do2017plant, lee2018multi}. Typically, in these models, latent representations are extracted from each image and fused into a single representation, which is then further processed by nonlinear layers \cite{seeland2021multi}. As a result, image-specific information is lost after fusion. In contrast, CST encoders are  specifically designed to integrate contextual information from multiple input images while preserving, for each image, individual representations that capture image-level peculiarities.

%% file: sec/4_proposed_solution.tex
\vspace{-0.3cm}
\section{Convolutional Set Transformer (CST)} \label{sec:cst_macrosection}
In this section, we introduce SetConv2D, the core building block of our architecture  (Section \ref{sec:setconv2d}), describe how to assemble CST networks (Section \ref{sec:cst_architecture}), and discuss CST explainability (Section \ref{sec:cst_explainability}).

\vspace{-0.1cm}
\subsection{SetConv2D} \label{sec:setconv2d}

SetConv2D is a convolutional block designed for processing arbitrarily sized sets of 3D volumes, which can be either input images or convolutional activation maps. SetConv2D operates as an equivariant set-to-set layer, \mbox{returning} an output volume for each input volume. Thus, SetConv2D retains a different representation for each image in the input set. More specifically, SetConv2D processes each volume within a set by preserving and exploiting the locality and translation invariance priors of CNNs, while leveraging the context information from other volumes within the set to enhance the output representation. The five stages composing SetConv2D are illustrated in Figure \ref{fig:setconv2d} and detailed in what follows.

\begin{itemize}
    \item[1)] \textbf{2D convolutional layer}. A shared convolutional layer processes each volume in the input set. Its hyperparameters—number of filters, \mbox{kernel} size, padding, and stride—are defined at SetConv2D initialization. \mbox{No activation} is applied at this stage.
    \item[2)] \textbf{Global Average Pooling}. Each volume returned by the convolutional layer at the previous step is reduced via GAP, yielding a latent vector. 
    \item[3)] \textbf{Multi-Head Self-Attention}. Latent vectors corresponding to different volumes in the set interact within a Multi-Head Self-Attention \mbox{module}, implemented as in \cite{vaswani2017attention}. Positional encoding is omitted to maintain permutation-equivariance.
    \item[4)] \textbf{Context-aware biases}. Thanks to skip connections illustrated in Figure \ref{fig:setconv2d}, each contextualized latent vector serves as a bias vector to be added to the corresponding volume computed at step 1).
    \item[5)] \textbf{Activation}. The volumes are finally passed through a non-linear \mbox{activation} function.
\end{itemize}

SetConv2D supports \texttt{valid} and \texttt{same} padding, stride $\geq 1$, and dilation. In our preliminary tests, we evaluated alternative pooling and attention mechanisms within SetConv2D blocks, and we observed that the combination of GAP and MHSA consistently delivered the strongest performance.

\subsection{CST Architecture}
\label{sec:cst_architecture}

CST consists of a permutation-equivariant encoder $\boldsymbol{\mathcal{E}}$, followed by a task-specific downstream network $\boldsymbol{\mathcal{H}}$, which may be either equivariant or invariant depending on the task. 

The encoder $\boldsymbol{\mathcal{E}}$ simultaneously performs feature extraction for each image in the input set while modeling the contextual relationships that arise within the set. It is built by stacking layers of SetConv2D blocks, much like Conv2D layers are stacked in standard CNNs. Since SetConv2D operates on arbitrarily sized sets and is permutation equivariant, the encoder naturally inherits these properties. Given an input image set, $\boldsymbol{\mathcal{E}}$ produces a set of contextualized latent representations:
\begin{equation}
\{R_1, \dots, R_N\} = \boldsymbol{\mathcal{E}}(\{I_1, \dots, I_N\})
\end{equation}

The downstream network $\boldsymbol{\mathcal{H}}$ further processes \mbox{$\{R_1, \dots, R_N\}$} to accomplish the specific task. For set-to-set tasks, $\boldsymbol{\mathcal{H}}$ operates on each individual image representation independently. In its simplest form, it can be a classifier or regressor that performs inference on each image's contextualized representation. On the other hand, for set-to-global tasks, $\boldsymbol{\mathcal{H}}$ typically employs an invariant pooling operation, such as summation or maximum, to aggregate \mbox{$\{R_1, \dots, R_N\}$} into a global set-level representation, which is further processed to produce a set-level prediction.

Similar to CNNs, CSTs extract hierarchical features by applying a series of convolution operations. The first SetConv2D layers react to fine-grained image patterns, such as edges and small details, while deep SetConv2D layers capture higher-level concepts. The key difference between a shared CNN backbone applied in parallel to each image in the input set and a CST encoder is that the former relies on static biases learned during training, whereas the latter, through the SetConv2D mechanism, employs dynamic biases that are adjusted at both training and inference time based on contextual information.

Note that CSTs can also process single images as sets of unit size. SetConv2D blocks are compatible with Conv2D layers and it is possible to interleave the two when designing CSTs, as we show in our experiments. This also implies that existing CNNs can be modified to include SetConv2D blocks, enabling training and inference on sets of images.

\subsection{Explainability}
\label{sec:cst_explainability}

Explainability in set-input networks is a largely unexplored research area. CST is the first architecture to provide demonstrated explainability support, as shown by our experiments, whereas competing approaches, such as Deep Sets and Set Transformer, remain inherently opaque \mbox{due to their design choices}.

As discussed in the previous sections, CSTs are constructed by stacking SetConv2D layers, in the same way that CNNs are built by stacking standard Conv2D layers. This design makes CST natively compatible with standard CNN explainability tools such as Grad-CAM. Since Grad-CAM relies on gradients with respect to the spatial representations produced by convolutional layers, it can be applied to the outputs of SetConv2D blocks, which provide a contextualized spatial representation for each image in the input set. The resulting explanation maps not only highlight which features of an image drive the prediction, but also reveal how these features are modulated by the presence of other images in the set. Notably, the same image can yield different explanation maps depending on the set it belongs to, making CST’s contextual reasoning directly observable.

Unlike CST, which natively operates on sets of 3D tensors such as images, Deep Sets and Set Transformer can only process vector embeddings. Consequently, images must first be reduced by a CNN to latent vectors, discarding spatial information before contextualization. Although Grad-CAMs can be computed with respect to the convolutional outputs of the CNN, this stage lies entirely before the Deep Sets or Set Transformer module. As a result, the explanation maps highlight features of individual images but offer no insight into the contextual modeling performed by the set-input network, which remains unexplained.

\section{CST Pre-training and Transfer Learning}
\label{sec:cst_pretraining}

Deep Sets and Set Transformers are usually trained from scratch, as in \cite{collenne2024reset}. Pre-trained set-learning models are not available, and Transfer Learning on these architectures has never been studied in the literature. In contrast, CST supports Transfer Learning in a natural way. Indeed, after pre-training on a large scale dataset like ImageNet, a CST encoder $\boldsymbol{\mathcal{E}}$ can be reused as a generic backbone for set-learning tasks. To adapt a pre-trained CST to a new task, the downstream network $\boldsymbol{\mathcal{H}}$ is re-initialized or even replaced with an architecture tailored to the task, while the encoder $\boldsymbol{\mathcal{E}}$ is either frozen or fine-tuned. To facilitate future research on Transfer Learning with CSTs, we publicly release CST-15, the first set-input backbone pre-trained on ImageNet.

In most real-world scenarios, training datasets provide only a single image per instance. This makes it infeasible to train set-input architectures such as Deep Sets, Set Transformers, or CSTs from scratch, since these models cannot learn how to leverage contextual interactions among images in a set, if only singleton sets are provided during training. CST addresses this challenge by enabling \textbf{Set-free Transfer Learning}: our experiments show that CST retains the ability to perform inference on arbitrarily large image sets with improved accuracy, \textit{even when the Transfer Learning phase is limited to training on individual images rather then image sets}. This is possible because the CST encoder $\boldsymbol{\mathcal{E}}$ preserves the contextualization capabilities it acquired during large-scale pre-training, thereby eliminating the need to re-learn how to exploit inter-image relationships when adapting to new tasks.

In Section \ref{sec:cic}, we propose Contextualized Image Classification as a pre-training objective for CSTs. In Section \ref{sec:CT}, we introduce Combinatorial Training, a training strategy that is highly beneficial when pre-training CSTs for the CIC task.

\subsection{Pre-training Objective: Contextualized Image Classification (CIC)}
\label{sec:cic}

We propose Contextualized Image Classification (CIC): a pre-training task for CST encoders, just like Image Classification is used to pre-train CNN encoders. In CIC, a permutation-equivariant network takes as input a set of images that all belong to the same (unknown) class and therefore share the same label. The model returns a \textit{contextualized prediction} for each image, based not only on the information specific to that image, but also on the contextual information provided by the other images in the set. By leveraging the shared context among the images, the network is able to improve its image-level classification accuracy.

We design a network for CIC pre-training by concatenating a CST encoder $\boldsymbol{\mathcal{E}}$ and a MLP classifier $\boldsymbol{\mathcal{H}}$. The encoder consists of a stack of SetConv2D blocks followed by GAP to collapse the spatial dimensions of the convolutional volumes returned by the last SetConv2D block. Thus, $\boldsymbol{\mathcal{E}}$ receives a set of images as input and returns a set of contextualized, image-level latent vectors. Each contextualized latent vector is then classified identically and independently by $\boldsymbol{\mathcal{H}}$. By training for CIC on large datasets, the encoder learns both to extract features from each image and to model the contextual relationships between different images in a set. After training, $\boldsymbol{\mathcal{H}}$ can be removed, and $\boldsymbol{\mathcal{E}}$ adapted to different set-learning
tasks, following standard Transfer Learning practice.

At first glance, the CIC task may seem counter-intuitive. Since all images in a set share the same label, it might appear more natural to aggregate their representations—e.g., by averaging—and produce a single set-level prediction, as in Set-level Classification. However, this yields a permutation-invariant network, which is unsuitable as a general-purpose encoder: set-pooling irreversibly collapses image-level representations that are crucial for set-to-set tasks such as Set Anomaly Detection. Moreover, removing the pooling module after training to recover equivariance is not viable, as it might disrupt the functioning of the network. In contrast, CIC enables the pre-training of permutation-equivariant CST encoders that preserve contextualized, image-level representations and can be seamlessly adapted to both set-to-set tasks, which require equivariant outputs, and set-to-global tasks, where invariance is readily obtained by applying a pooling operation to the outputs of the equivariant backbone $\boldsymbol{\mathcal{E}}$. In practice, CIC also yields more stable training than Set-level Classification.

Importantly, the same dataset can be annotated at different levels of granularity, giving us an additional degree of freedom to exploit during CST pre-training with CIC. For instance, ImageNet \cite{deng2009imagenet} organizes an ontology of images based on the semantic hierarchy of WordNet \cite{fellbaum1998wordnet}. Depending on the selected level of granularity, an image of an husky can be assigned to the class \textit{husky}, \textit{working dog}, \textit{dog}, \textit{canine}, \textit{carnivore}, \textit{placental}, or \textit{mammal}. Higher granularity results in more classes and lower intra-class variability, while lower granularity leads to fewer, coarser classes with greater intra-class variability. By selecting the granularity of the labels, we can influence the type of features that the CST learns during pre-training. Finer granularity encourages the network to capture subtle distinctions between visually similar categories, leading to more specialized representations. Coarser granularity, instead, pushes the model to focus on higher-level semantic connections between images within a set, resulting in more general representations.

\subsection{Combinatorial Training (CT)}\label{sec:CT}

In the CIC task, a network predicts the class of an image conditionally on a variable number of other images from the same (unknown) class. As a result, when training a model for CIC, the number of training samples scales combinatorially with the number of possible sets that can be formed from images of the same class. We leverage this insight to design Combinatorial Training (CT), a novel training strategy for CST architectures and, in principle, for any set-input network.

In CT, before each epoch starts, we draw a set size $n$ uniformly at random in the range $[n_{\min}, n_{\max}]$, where $n_{\min}$ and $n_{\max}$ are hyperparameters controlling the minimum and maximum number of images in a set. Then, training images within each class are randomly assembled into sets of size $n$, and sets from all the classes are randomly assigned to batches of size $b$. The batch size $b$ defines the number of sets in a batch, which is kept fixed across epochs. Consequently, since the number of images per set varies in $[n_{\min}, n_{\max}]$, the total number of images per batch varies in $[b \times n_{\min}, b \times n_{\max}]$ during training.

CT operates as an effective form of data augmentation, since the chances of the model encountering the same set more than once during training are extremely low. Furthremore, concurrent training on sets of varying sizes allows the model to learn how to handle input sets of different sizes effectively. Importantly, the benefits of CT are not due to permuting the elements of the set—since CSTs are either permutation-equivariant or invariant—but rather stem from building random image sets of varying size during training, which has a data augmentation effect. Setting $n_{min}=1$ can speed up convergence when training on large datasets, as training only with sets may act as an overly strong augmentation that slows optimization.

CT enables memory-efficient training of Convolutional Set Transformers as small sets can be used for training, while preserving the model's capacity to handle arbitrarily large sets during inference. This strategy ensures scalable performance without incurring the high memory costs typically associated with larger sets during training. For example, the CST-15 model is trained on ImageNet using a minimum set size of $n_{\text{min}} = 1$ and a maximum set size of $n_{\text{max}} = 2$, i.e., it is trained on individual images (as in standard CNN training) and pairs of images from the same class. Despite this, at inference time, it can process image sets of any size with increasing accuracy, as demonstrated in Section \ref{sec:cst15}.

%% file: sec/5_experiments.tex
\section{Experiments}
\label{sec:experiments}

In Section \ref{sec:tiny_experiments}, we show that CST outperforms Deep Sets and Set Transformer in Set Classification tasks across several datasets. In Section \ref{sec:anomaly} we focus on Set Anomaly Detection and demonstrate that CST achieves superior performance while enabling explainability. Finally, in Section \ref{sec:cst15}, we introduce CST-15, a CST pre-trained on ImageNet, showing that it can be adapted to new tasks by Transfer Learning.

\subsection{Classification Experiments}
\label{sec:tiny_experiments}

We carried out extensive experiments to benchamrk CSTs against Deep Sets, Set Transformers, and CNNs in the Contextualized Image Classification (CIC) and Set-level Classification (SC) tasks. CIC is introduced in Section \ref{sec:cic} as a pre-training task, but we can also test on this as the ultimate network task. In both CIC and SC, a network receives as input a set of images belonging to the same unknown class, but in SC the network predicts a single class probability distribution at the set level, while in CIC the output consists of multiple (contextualized) distributions at the image level. Notably, CIC models are by design equivariant with respect to permutations of the input set, while SC models are invariant.

\subsubsection{Tested Architectures}
\label{sec:tested_architectures}

We tested many CIC and SC models across several datasets, including ImageNet64x64\footnote{In this section, we use ImageNet64x64, a downsampled version of ImageNet, to keep computations tractable when comparing architectures. In Section \ref{sec:cst15}, we will present CST-15, a CST pre-trained on the ImageNet dataset at standard $224 \times 224$ resolution.} \cite{chrabaszcz2017downsampled}, Tiny ImageNet \cite{le2015tiny}, CIFAR-10, and CIFAR-100 \cite{krizhevsky2009learning}, showing that CSTs consistently outperform baselines models, often by a substantial margin. For each dataset, we trained the following \textbf{equivariant networks for Contextualized Image Classification} (see also Table \ref{tab:architecture_overview}):
\begin{itemize}
    \item A CST designed by stacking pairs of SetConv2D blocks interleaved with max-pooling layers. For models trained on ImageNet64x64 and Tiny ImageNet, the the architecture comprises $8$ SetConv2D blocks, whereas for the CIFAR datasets it uses $6$. Each SetConv2D has a kernel size of $3$ with \texttt{same} padding, and the number of attention heads is set to $min(n_{filters}, 64)$. The final SetConv2D block is followed by GAP and a dense layer with softmax activation to produce the class probabilities.
\input{tables/architectures.tex}
    \item A CNN obtained from the CST by replacing all SetConv2D blocks with standard Conv2D layers. We refer to this CNN as the \textit{equivalent CNN} since it preserves the same convolutional structure and receptive field as the corresponding CST. Unlike the CST model, however, the \textit{equivalent CNN} does not support set processing and can operate only on individual images. Table \ref{tab:architecture_overview} reports a higher parameter count for CST compared to the \textit{equivalent CNN}, but this difference is not substantial, as the two architectures are directly comparable only when operating on individual images (set size $=1$). In this setting, the MHSA component within each SetConv2D block collapses to a single linear channel projection, so CST and its \textit{equivalent CNN} present the same number of nonlinearities. Furthermore, CST still retains the full set of MHSA weight matrices within each SetConv2D block—including the now functionally redundant query and key matrices—which inflates its reported parameter count.
    \item Two Set Transformers \cite{lee2019set} with two (ST-S) and three (ST-L) Set Attention Blocks (SABs) on the top of the \textit{equivalent CNN}. ST-S has fewer parameters than the competing CST, while ST-L has more parameters. A GAP layer is applied to each convolutional volume returned by the \textit{equivalent CNN}, since the subsequent Set Attention Blocks are designed to operate on sets of vectors rather than spatial tensors. As in \cite{lee2019set}, Set Attention Blocks are defined as an adaptation of the encoder block of the Transformer \cite{vaswani2017attention}, without positional encoding and dropout:
    \begin{equation}
    SAB(S_N) = LayerNorm(H+rFF(H))
    \end{equation}
    \begin{equation}
        H=LayerNorm(S_N+MHSA(S_N))
    \end{equation}
   Here, $S_N \in \mathbb{R}^{N \times C}$ represents the input set, with rows corresponding to latent vectors. The notation $MHSA(\cdot)$ denotes Multi-Head Self-Attention, while $rFF(\cdot)$ refers to a row-wise feedforward layer that processes all the latent vectors independently.
    \item A Deep Sets \cite{zaheer2017deep} network (DS), with three Deep Sets equivariant layers on the top of the \textit{equivalent CNN}. As for ST models, GAP is applied to each convolutional volume computed by the \textit{equivalent CNN}, yielding a set of latent vectors, since Deep Sets layers cannot handle sets of spatial tensors. We investigated several implementations of the Deep Sets layers. In the best-performing variant, the mean latent vector of the set is added to each individual vector, and the result is then passed through a dense layer with nonlinear activation:
    \begin{equation}
    DeepSets(S_N) = rFF(S_N + \frac{1}{N}\textbf{1}\textbf{1}^TS_N)
    \end{equation}
    Here, $S_N \in \mathbb{R}^{N \times C}$  represents the input set, $\textbf{1} \in \mathbb{R}^{N}$, and $rFF(.)$ \mbox{denotes} a row-wise feedforward layer that processes latent vectors independently and identically. 
    In our experiments, increasing the number of Deep Sets layers beyond three leads to lower performance. 
\end{itemize}

Furthermore, for each dataset, we assessed the following \textbf{invariant architectures for Set-level Classification}:
\begin{itemize}
    \item  Score Fusion (SF) models \cite{seeland2021multi}. By default, the CIC models reported in Table \ref{tab:architecture_overview} output image-level predictions, but can be adapted to generate set-level predictions without any further training. Indeed, at inference time, we can simply average the estimated class probabilities from all the images, yielding a set-level distribution.
    \item Late Fusion (LF) models \cite{seeland2021multi}. An invariant set pooling module can be added to the CIC architectures of Table \ref{tab:architecture_overview}, placed just before the final classifier, enabling set-level predictions. The pooling module collapses the image-level latent vectors into an aggregate set-level representation, before classification. In our implementation, we compute the mean latent vector of the set and feed it to a dense layer with a non-linear activation:
    \begin{equation}
        LateFusion(S_N) = \sigma(\beta + \frac{1}{N}\textbf{1}^TS_N\Gamma)
    \end{equation}
     Here, $S_N \in \mathbb{R}^{N \times C}$  represents the input set,  $\Gamma$ and $\beta$ are learnable weights and biases, $\sigma$ is a non-linear activation function, while $\textbf{1} \in \mathbb{R}^{N}$. In all Late Fusion models, we introduce a Layer Normalization \cite{ba2016layer} step following the fusion module, as it enhances training stability, particularly with complex datasets such as ImageNet64x64. We train Late Fusion networks from scratch, separately from the corresponding CIC models.
\end{itemize}

\subsubsection{Training Hyperparameters}
\label{sec:combinatorial_training}

We trained both CSTs and all the other models---except for standard CNNs, which completely disregard set information---using Combinatorial Training with $n_{min}=2$ and $n_{max}=5$. Note that CT, introduced in Section \ref{sec:CT} as a training strategy to be used in conjunction with CIC, can be applied to SC as well. 

\begin{figure}[ht]
    \centering
    \includegraphics[width=0.60\textwidth]{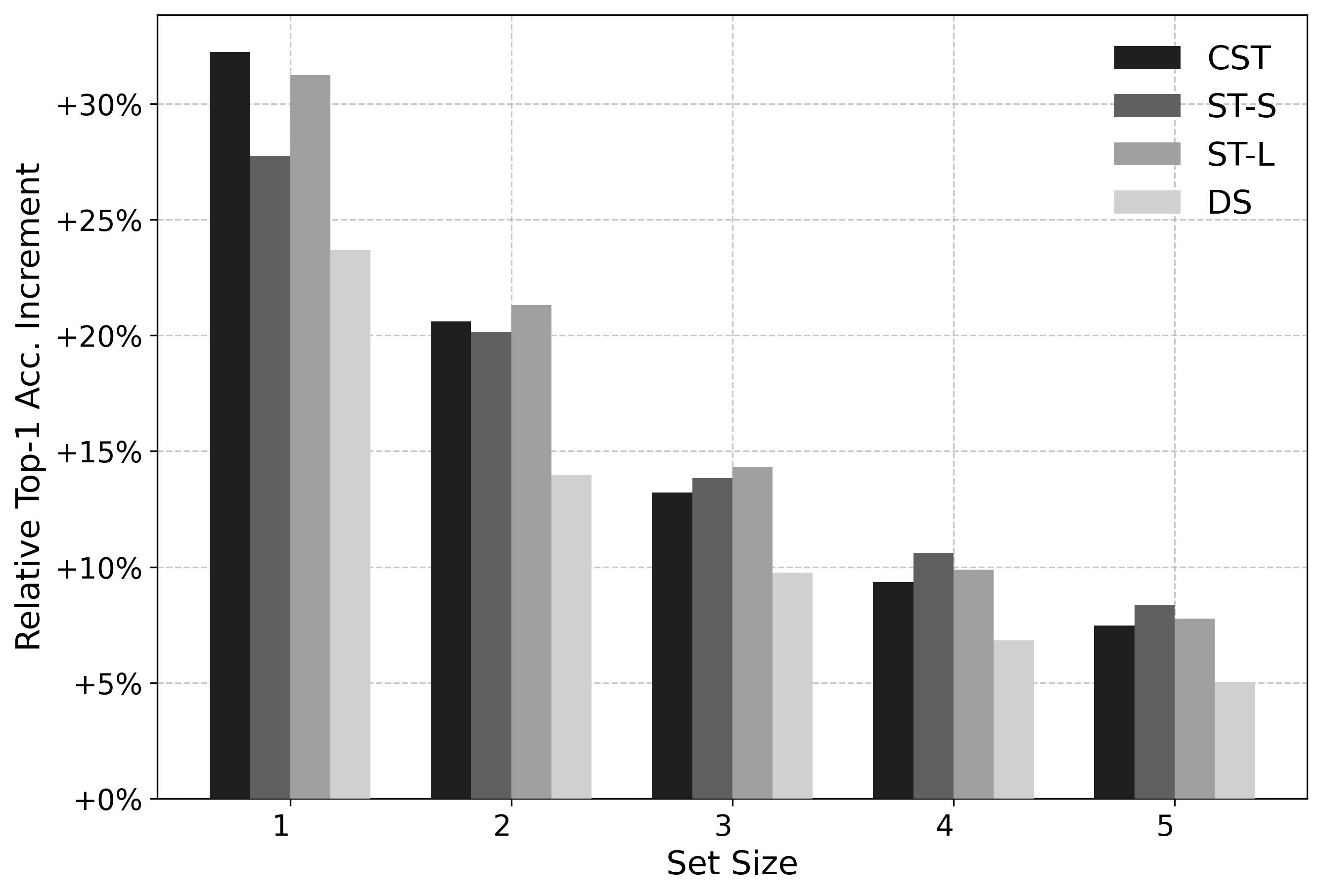}
    \caption{\textbf{Combinatorial Training ablation}. Relative increase in Top-1 Test Accuracy for CIC models trained on the ImageNet64x64 dataset when employing CT compared to conventional training.}
    \label{fig:ct_ablation}
\end{figure}

Our analysis reveals that CT benefits both CSTs and baseline models. Figure \ref{fig:ct_ablation} shows the relative increase in Top-1 test accuracy for CIC models trained on the ImageNet64x64 dataset when employing CT compared to conventional training\footnote{When CT is not applied, we create sets of three images from the same class \textit{before the training process begins} and use these sets to train the model in every epoch.}. CT is advantageous for CSTs, Set Transformers, and Deep Sets models, with relative improvements in Top-1 accuracy ranging from +24\% to +32\% for sets of size 1, and from +14\% to +21\% for size 2. We report an in-depth ablation study on Combinatorial Training in Appendix \ref{sec:ablation_ct}, which confirms these findings.

To ensure comparability, we trained all competing models using the same hyperparameters and fixed random seeds. Our results are robust across different hyperparameter configurations. A detailed description of the hyperparameters used can be found in Appendix \ref{sec:experiment_details}.

\subsubsection{Results and Analysis}\label{sec:results}

Table \ref{tab:performance_summary} shows the Top-1 test accuracy as a function of the input set size for CIC and SC tasks across all models trained on ImageNet64x64, Tiny ImageNet, CIFAR-10, and CIFAR-100. CSTs outperform the baselines in 58 out of 60 cases, spanning various combinations of dataset, task, fusion strategy (for SC models), and input-set size. In \textit{relative terms}, CSTs outperform the best alternative up to +20.3\%, and by more than +10\% in 17 cases. 

For the sake of convenience, we report results for sets of up to five images; however, all set-based models can process arbitrarily large image collections, with performance improvements as the set size grows. As expected, performance exhibits diminishing returns with increasing set size: for larger input sets, additional context images provide mostly redundant information, and performance gains saturate. Moreover, the accuracy of different architectures converges as the context size increases, suggesting that the model choice becomes less critical when the context is sufficiently large.

\begin{figure*}[!t]
    \centering
    \includegraphics[width=\textwidth]{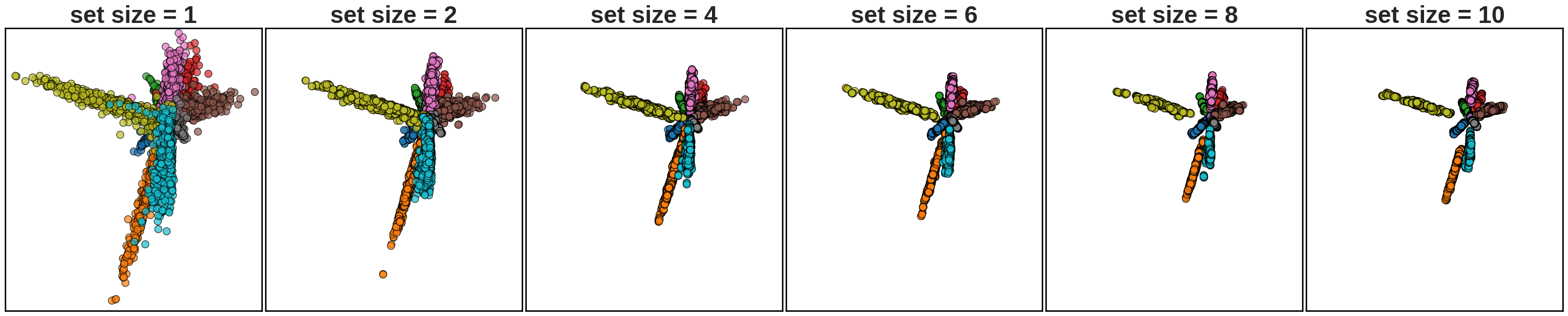}
    \caption{\textbf{2D latent representations of CIFAR-10 test images with varying context size}. Each color corresponds to a class.}
    \label{fig:latent_space}
\end{figure*}

To study how larger context sharpens class separation in the latent space of a CST trained for CIC, we designed a controlled experiment on CIFAR-10. We trained a modified version of the CIFAR-10 CST model presented in Table \ref{tab:architecture_overview}, inserting a bottleneck—a linear layer with two neurons and no activation—after GAP and before the classifier. Once trained, we removed the classifier, leaving a CST encoder that outputs a contextualized 2D embedding for each image in a set. Using this encoder, we mapped each CIFAR-10 test image into the latent space by placing it within sets of $N = 1,2,4,6,8,10$ images sampled from the same class. Figure \ref{fig:latent_space} illustrates the resulting embeddings across different set sizes. When no context is available ($N=1$), embeddings are scattered throughout the latent space, with substantial overlap across classes. As context increases, the model progressively refines the embeddings, leading to clearer separation. With sufficient context ($N=10$), the ten classes emerge as well-separated clusters with minimal overlap.

\input{tables/performance.tex}

\vspace{-0.5cm}

\vspace{0.6cm}
\subsection{Explainable Set Anomaly Detection}
\label{sec:anomaly}

Set-input networks like Deep Sets and Set Transformer have been applied to Set Anomaly Detection \cite{zaheer2017deep, lee2019set}, a task where a model receives a set of images and must identify any anomalous image. Anomalous images stand out within a set by either lacking an element that is present in most of the other images or including an element that is absent from the majority. While Deep Sets and Set Transformers can identify such anomalous images, they fail to explain why an image is considered anomalous, highlighting image regions that influenced classification. In this experiment, we demonstrate that CSTs outperform Deep Sets and Set Transformers in Set Anomaly Detection and also offer reliable explanations of model predictions.

CelebA \cite{liu2015faceattributes} is a dataset containing 202,599 facial images across 10,177 subjects. Each image is annotated with 40 binary attributes, which describe physical characteristics (e.g., oval face), facial expressions (e.g., smiling), and the presence of accessories (e.g., eyeglasses). We utilize CelebA attributes to form image sets. Each image set is obtained by selecting a majority of images that exhibit two randomly chosen attributes (normal images), along with a minority of images in which both these attributes are absent (anomalous images). The total number of images in a set, as well as the percentage of anomalous images, varies across sets. Sets without any anomalies are also allowed. We train a CST network, two Set Transformers (ST-S and ST-L), and a Deep Sets network (DS) to predict, for each image in a set, whether it is anomalous or not with respect to the set. This corresponds to a binary classification task. The experiment is conducted at the original CelebA resolution of $178 \times 218$ pixels, without any cropping. Details on the tested architectures and the training hyperparameters can be found in Appendix \ref{sec:experiment_details}.

\begin{table}[h]
    \centering
    \resizebox{0.60\textwidth}{!}{
    \begin{tabular}{rr|cccc}
        \toprule
        \multirow{3}{*}{\textbf{Anomaly \%}} & \multirow{3}{*}{\textbf{Set size}} & \textbf{CST} & \textbf{ST-S} & \textbf{ST-L} & \textbf{DS} \\
        & & 27.8M & 27.3M & 28.3M & 19.1M \\
        & & params. & params. & params. & params. \\
        \midrule
        \multirow{3}{*}{0.1} & 10 & \textbf{0.8133} & 0.7878 & 0.8052 & 0.7273 \\
        & 20 & \textbf{0.8300} & 0.8105 & 0.8252 & 0.7456 \\
        & 40 & \textbf{0.8326} & 0.8188 & 0.8302 & 0.7505 \\
        \midrule
        \multirow{3}{*}{0.2} & 10 & \textbf{0.8886} & 0.8791 & 0.8880 & 0.8160 \\
        & 20 & \textbf{0.8966} & 0.8843 & 0.8948 & 0.8286 \\
        & 40 & \textbf{0.8989} & 0.8896 & 0.8970 & 0.8363 \\
        \midrule
        \multirow{3}{*}{0.3} & 10 & \textbf{0.9182} & 0.9158 & 0.9141 & 0.8455 \\
        & 20 & \textbf{0.9261} & 0.9215 & 0.9215 & 0.8662 \\
        & 40 & \textbf{0.9272} & 0.9231 & 0.9221 & 0.8740 \\
        \midrule
        \multirow{3}{*}{0.4} & 10 & \textbf{0.8991} & 0.8891 & 0.8918 & 0.8535 \\
        & 20 & \textbf{0.9170} & 0.9084 & 0.9122 & 0.8760 \\
        & 40 & \textbf{0.9224} & 0.9152 & 0.9185 & 0.8866 \\
        \bottomrule
    \end{tabular}}
    \caption{\textbf{Set Anomaly Detection}. AUPRC at various input set sizes and anomaly prevalence rates. DS has fewer parameters than other architectures as larger DS models (with more Deep Sets layers) exhibit significant convergence issues.}
    \label{tab:anomaly_detection}
\end{table}

Results, presented in Table \ref{tab:anomaly_detection}, demonstrate that CSTs consistently outperform competing models across various set sizes and anomaly percentages. Furthermore, CSTs can generate interpretable explanations for model predictions. Figure \ref{fig:gradcam_anomaly} illustrates some image sets with GradCAMs overlaid on anomalous images, providing visualizations for both CST and ST-L. CST explanations are intuitive and meaningful. For example, in the first set, attention is drawn to the head and mouth, as anomalous images represent individuals without a hat and not smiling. On the other hand, ST-L Grad-CAMs fail to provide meaningful explanations. This is not unexpected, given that calculating Grad-CAMs for Set Transformer involves linearizing the entire stack of Set Attention Blocks, which is the network component responsible for modeling the contextual relationships between set images. In contrast, CSTs include only one linear layer after the last SetConv2D block as the contextual relationships between images in a set are fully modeled in the stack of SetConv2D blocks, jointly with feature extraction.

\subsection{Pre-training CSTs for Transfer Learning}
\label{sec:cst15}

In the image processing domain, Deep Sets and Set Transformers are typically trained from scratch, as in \cite{collenne2024reset}. Pre-trained models are not available, and Transfer Learning on these architectures has not been explored yet. In this section, we demonstrate that CSTs can be pre-trained on large-scale datasets and adapted to new tasks in Transfer Learning scenarios. To support this claim, we introduce and publicly release for download CST-15\footnote{\url{https://github.com/chinefed/convolutional-set-transformer}}, a Convolutional Set Transformer pre-trained on the ImageNet dataset \cite{russakovsky2015imagenet} for the Contextualized Image Classification task. Section \ref{sec:cst15_pretraining} provides an overview of the CST-15 architecture and its performance on the ImageNet validation set, while Section \ref{sec:pec_tl} presents the results of a challenging Transfer Learning task: event recognition from personal photo albums.

\vspace{-0.10cm}
\subsubsection{CST-15 Trained on ImageNet}
\label{sec:cst15_pretraining}

\input{tables/cst_for_tl.tex}

Table \ref{tab:cst15} summarizes the CST-15 architecture and compares it to VGG-19 \cite{simonyan15very}, which we selected as a reference due to its similar convolutional structure. Unlike VGG-19, which relies exclusively on standard convolutions, CST-15 employs SetConv2D blocks, allowing for concurrent context modeling and feature extraction. Moreover, CST-15 employs Global Average Pooling followed by a single classification layer, instead of the flattening and fully connected layers used in VGG-19, resulting in a significantly smaller model size (28M vs. 144M parameters). Finally, CST-15 uses ReLU6 activations \cite{krizhevsky2010convolutional}, which improve robustness in low-precision environments by capping activation values at 6 \cite{howard2017mobilenets, sandler2018mobilenetv2}.

We train CST-15 from scratch for the CIC task,
using Combinatorial Training at $224 \times 224$ resolution. The CT hyperparameters $n_{min}$ and $n_{max}$ are set to 1 and 2, respectively. This setup accelerates convergence on a complex dataset like ImageNet and reduces the memory footprint, as the model is trained only on small sets—single images and pairs of images—yet it still generalizes to arbitrarily large sets at inference time. We refer to Appendix \ref{sec:experiment_details} for a detailed description of the training setup.

Table \ref{tab:cst15} compares the CIC performance of CST-15 on the ImageNet validation set with the single-image classification performance of VGG-19\footnote{We obtain the VGG-19 ImageNet weights from TensorFlow.}. For a set size of $1$, CIC reduces to standard single-image classification, where CST-15 already outperforms VGG-19, despite the lower parameter count. Furthermore, while VGG-19 operates only on individual images, CST-15 can handle arbitrarily large input sets. By exploiting contextual relationships among images in a set, CST-15 achieves consistent performance gains as the set size increases.

As discussed in Section \ref{sec:anomaly}, unlike Deep Sets and Set Transformers, CSTs are readily compatible with standard explainability tools used for CNNs. Appendix \ref{sec:explain_cst} presents a qualitative evaluation of CST-15's explainability on the ImageNet dataset.

\subsubsection{Transfer Learning on the PEC dataset}
\label{sec:pec_tl}

\begin{figure*}[ht]
    \centering
    \includegraphics[width=0.6\textwidth]{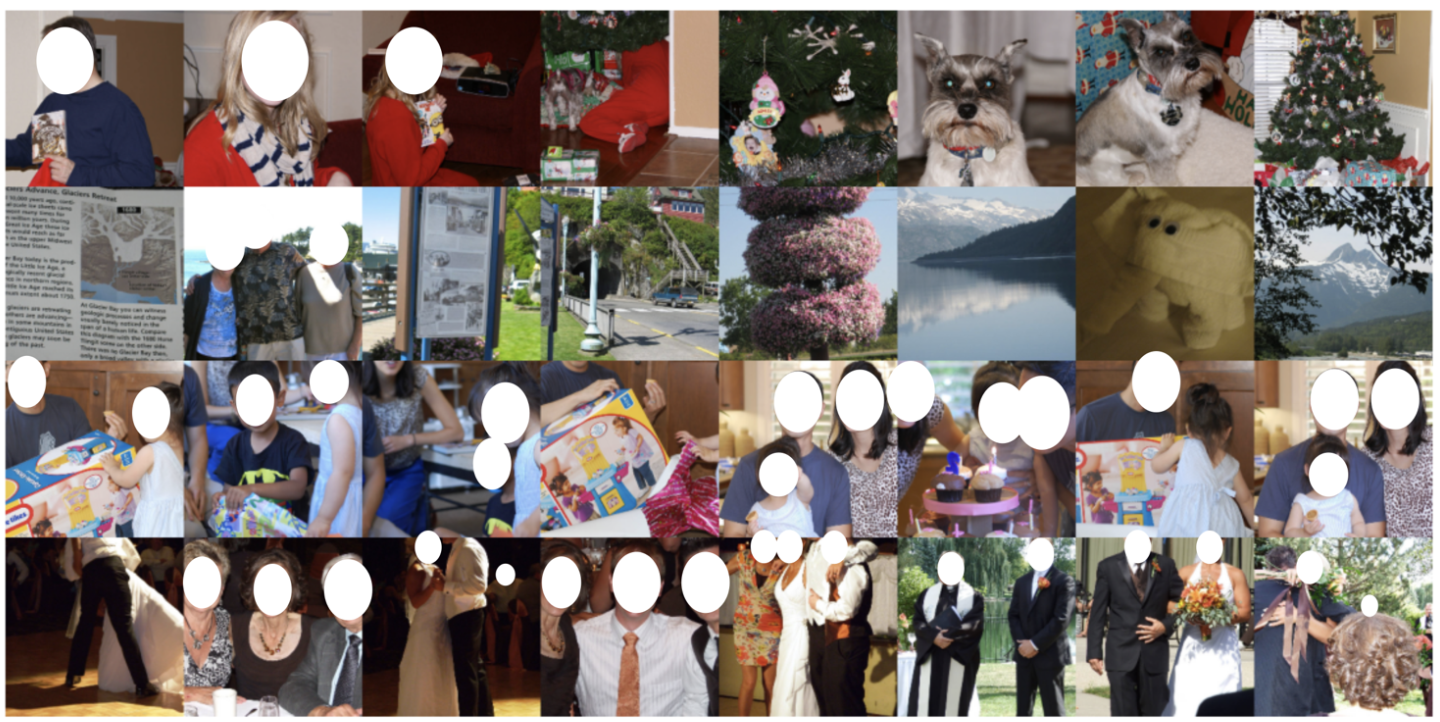}
    \caption{\textbf{Four image sets drawn from the PEC dataset.} Each row corresponds to an image set, with all images in a set drawn from the same personal album. From top to bottom, the event categories are Christmas, Cruise, Children’s Birthday, and Wedding. Notably, some images are irrelevant to the respective event category, introducing noise. In the figure, faces are masked to comply with the dataset’s usage policies, although the experiment was conducted on the original, unmasked images.}
    \label{fig:pec}
\end{figure*}

In this section, we show that CSTs pre-trained on large-scale datasets can be adapted to new set-learning tasks via standard Transfer Learning, similarly to CNNs. As a case study, we consider event recognition from personal photos, where images are grouped into albums documenting individual events. For this purpose, we use the PEC dataset \cite{bossard13} which includes 807 personal photo albums, totaling 61,364 images. Each album, captured by a single photographer, documents a specific event and is labeled with one of 14 categories such as birthday, concert, or road trip. All images are center-cropped and resized to $224 \times 224$ pixels. Notably, PEC is a challenging dataset, often containing cluttered or irrelevant images to event recognition, such as accidental shots, close-ups of individuals, and photos of random objects. Figure \ref{fig:pec} shows some image sets drawn from the PEC dataset.

We assess a CIC task, where each individual photo must be assigned the event label of its album based both on its own content and on contextual information from the other photos in the album. This formulation is substantially more challenging than album-level classification which, like SC, collapses all image representations into a single album-level representation, loosing photo-specific information prior to prediction.  By contrast, CIC forces the network to preserve distinct yet contextualized representations for each image, encoding the both the image’s unique content and the broader \mbox{album context}.

We compare the performance of our CST-15 model, the first set-input encoder pre-trained on ImageNet, with that of VGG-19 applied in parallel to each image in a set. For Transfer Learning, we adapt both CST-15 and VGG-19 by replacing their original 1000-way ImageNet classifier with a new layer for the 14 PEC classes, freezing the backbones and training only the final classifier. To ensure a fair comparison, CST-15 is trained using the Set-free Transfer Learning scheme introduced in Section \ref{sec:cst_pretraining}. Consequently, during the Transfer Learning phase, both CST-15 and VGG-19 are exposed only to individual images (i.e., singleton sets) without any context. Nevertheless, at inference time, CST-15 can process arbitrarily large sets of images from the same personal album, classifying each image with increasing accuracy as the set size grows, since it has learned to exploit contextual relationships among images during its ImageNet pre-training (see Figure \ref{fig:pec}). In contrast, VGG-19’s performance remains unaffected by the set size, since it does not support set-based inputs and thus can only process each image in a set independently, without leveraging contextual information from the other images. Notably, CST-15 outperforms VGG-19 even when the set size is $1$, i.e., when both models receive a single image as input without any context. Additional details about the experiment are available in \mbox{Appendix \ref{sec:experiment_details}}.

%% file: tables/architectures.tex
\begin{table*}[h]
    \centering
    \huge
    \resizebox{\textwidth}{!}{
\begin{tabular}{c|c|c|c|c|c|c|c|c|c|c|c|c|c|c}
    \toprule
    \multicolumn{5}{c|}{\textbf{ImageNet64x64}} & \multicolumn{5}{c|}{\textbf{Tiny ImageNet}} & \multicolumn{5}{c}{\textbf{CIFAR-10/100}} \\
    \midrule
    \textbf{CST} & \textbf{CNN} & \textbf{ST-S} & \textbf{ST-L} & \textbf{DS} & \textbf{CST} & \textbf{CNN} & \textbf{ST-S} & \textbf{ST-L} & \textbf{DS} & \textbf{CST} & \textbf{CNN} & \textbf{ST-S} & \textbf{ST-L} & \textbf{DS} \\
    \midrule
    8M & 5.2M & 7.3M & 8.4M & 6M & 1.9M & 1.2M & 1.8M & 2M & 1.6M & 462K/474K & 288K/300K & 421K/433K & 487K/499K & 387K/398K \\
    params. & params. & params. & params. & params. & params. & params. & params. & params. & params. & params. & params. & params. & params. & params. \\
    \midrule
    \multicolumn{5}{c|}{\textbf{Input: 64 $\times$ 64  RGB images}} & \multicolumn{5}{c|}{\textbf{Input: 64 $\times$ 64  RGB images}} & \multicolumn{5}{c}{\textbf{Input: 32 $\times$ 32  RGB images}}  \\
    \midrule
     SetConv2D(64, 3) & \multicolumn{4}{c|}{Conv2D(64, 3)} & SetConv2D(32, 3) & \multicolumn{4}{c|}{Conv2D(32, 3)} & SetConv2D(32, 3) & \multicolumn{4}{c}{Conv2D(32, 3)} \\
     SetConv2D(64, 3) & \multicolumn{4}{c|}{Conv2D(64, 3)} & SetConv2D(32, 3) & \multicolumn{4}{c|}{Conv2D(32, 3)} & SetConv2D(32, 3) & \multicolumn{4}{c}{Conv2D(32, 3)} \\
    \midrule
    \multicolumn{5}{c|}{\textbf{Maxpool}} & \multicolumn{5}{c|}{\textbf{Maxpool}} & \multicolumn{5}{c}{\textbf{Maxpool}} \\
    \midrule
    SetConv2D(128, 3) & \multicolumn{4}{c|}{Conv2D(128, 3)} & SetConv2D(64, 3) & \multicolumn{4}{c|}{Conv2D(64, 3)} & SetConv2D(64, 3) & \multicolumn{4}{c}{Conv2D(64, 3)} \\
    SetConv2D(128, 3) & \multicolumn{4}{c|}{Conv2D(128, 3)} & SetConv2D(64, 3) & \multicolumn{4}{c|}{Conv2D(64, 3)} & SetConv2D(64, 3) & \multicolumn{4}{c}{Conv2D(64, 3)} \\
    \midrule
    \multicolumn{5}{c|}{\textbf{Maxpool}} & \multicolumn{5}{c|}{\textbf{Maxpool}} & \multicolumn{5}{c}{\textbf{Maxpool}} \\
    \midrule
    SetConv2D(256, 3) & \multicolumn{4}{c|}{Conv2D(256, 3)} & SetConv2D(128, 3) & \multicolumn{4}{c|}{Conv2D(128, 3)} & SetConv2D(128, 3) & \multicolumn{4}{c}{Conv2D(128, 3)} \\
    SetConv2D(256, 3) & \multicolumn{4}{c|}{Conv2D(256, 3)} & SetConv2D(128, 3) & \multicolumn{4}{c|}{Conv2D(128, 3)} & SetConv2D(128, 3) & \multicolumn{4}{c}{Conv2D(128, 3)} \\
    \midrule
    \multicolumn{5}{c|}{\textbf{Maxpool}} & \multicolumn{5}{c|}{\textbf{Maxpool}} & \multicolumn{5}{c}{\textbf{Maxpool}} \\
    \midrule
    SetConv2D(512, 3) & \multicolumn{4}{c|}{Conv2D(512, 3)} & SetConv2D(256, 3) & \multicolumn{4}{c|}{Conv2D(256, 3)} & - & \multicolumn{4}{c}{-} \\
    SetConv2D(512, 3) & \multicolumn{4}{c|}{Conv2D(512, 3)} & SetConv2D(256, 3) & \multicolumn{4}{c|}{Conv2D(256, 3)} & - & \multicolumn{4}{c}{-} \\
    \midrule
    \multicolumn{5}{c|}{\textbf{Maxpool}} & \multicolumn{5}{c|}{\textbf{Maxpool}} & \multicolumn{5}{c}{-} \\
    \midrule
    \multicolumn{5}{c|}{\textbf{GAP}} & \multicolumn{5}{c|}{\textbf{GAP}} & \multicolumn{5}{c}{GAP} \\
    \midrule
    - & - & SAB(512) & SAB(512) & DS(512) & - & - & SAB(256) & SAB(256) & DS(256) & - & - & SAB(128) & SAB(128) & DS(128) \\
    - & - & SAB(512) & SAB(512) & DS(512) & - & - & SAB(256) & SAB(256) & DS(256) & - & - & SAB(128) & SAB(128) & DS(128) \\
    - & - & - & SAB(512) & DS(512) & - & - & - & SAB(256) & DS(256) & - & - & - & SAB(128) & DS(128) \\
    \midrule
    \multicolumn{5}{c|}{FC(1000)} & \multicolumn{5}{c|}{FC(200)} & \multicolumn{5}{c}{FC(10/100)} \\
    \midrule
    \multicolumn{5}{c|}{\textbf{Softmax}} & \multicolumn{5}{c|}{\textbf{Softmax}} & \multicolumn{5}{c}{\textbf{Softmax}} \\
    \bottomrule
\end{tabular}
    }
    \caption{\textbf{Overview of architectures trained for the Contextualized Image Classification (CIC) task}. SetConv2D blocks and Conv2D layers specified with ($n_{filters}$, kernel size), All convolutions are implemented with \texttt{same} padding. In SetConv2D blocks, attention heads have dimension $min(n_{filters}, 64)$. Maxpooling is implemented with pool size $2$. For Set Attention Blocks (SAB) and Deep Sets blocks (DS), we report the dimension of the output space in parenthesis. In all architectures, we use ReLU activations.}
    \label{tab:architecture_overview}
\end{table*}

%% file: tables/performance.tex
\begin{table*}[!t]
    \centering
    \huge
    \resizebox{\textwidth}{!}{ 
    \begin{tabular}{r | ccccc | ccccc | ccccc | ccccc}
        \toprule
        \textbf{Model} & \multicolumn{5}{c}{\textbf{ImageNet64x64}} & \multicolumn{5}{c}{\textbf{Tiny ImageNet}} & \multicolumn{5}{c}{\textbf{CIFAR 10}} & \multicolumn{5}{c}{\textbf{CIFAR 100}} \\
        \cmidrule(lr){2-6} \cmidrule(lr){7-11} \cmidrule(lr){12-16} \cmidrule(lr){17-21}
        & 1 & 2 & 3 & 4 & 5 & 1 & 2 & 3 & 4 & 5 & 1 & 2 & 3 & 4 & 5 & 1 & 2 & 3 & 4 & 5 \\
        \midrule
        \multicolumn{21}{c}{\textbf{Contextualized Image Classification (CIC)}} \\
        \midrule
        CST & 39.30 & \textbf{63.71} & \textbf{76.46} & \textbf{83.14} & \textbf{87.02} & \textbf{39.04} & \textbf{59.87} & \textbf{74.16} & \textbf{81.07} & \textbf{87.15} & \textbf{80.59} & \textbf{94.06} & \textbf{97.48} & \textbf{98.89} & \textbf{99.16} & \textbf{49.49} & \textbf{70.81} & \textbf{82.41} & \textbf{88.56} & \textbf{92.33} \\
        \textit{$\Delta$} & \textit{-2.28} & \textit{+3.91} & \textit{+2.93} & \textit{+1.72} & \textit{+0.85} & \textit{+5.51} & \textit{+8.06} & \textit{+8.57} & \textit{+4.78} & \textit{+6.37} & \textit{+1.20} & \textit{+1.19} & \textit{+0.45} & \textit{+0.55} & \textit{+0.23} & \textit{+5.87} & \textit{+7.38} & \textit{+6.99} & \textit{+5.22} & \textit{+4.33} \\
        CNN & \textbf{41.58} & - & - & - & - & 33.53 & - & - & - & - & 79.39 & - & - & - & - & 43.62 & - & - & - & - \\
        ST-S & 35.58 & 59.80 & 73.53 & 81.42 & 86.17 & 31.24 & 51.81 & 65.08 & 73.91 & 79.60 & 79.22 & 92.87 & 97.03 & 98.34 & 98.93 & 42.19 & 63.43 & 75.42 & 83.34 & 88.00 \\
        ST-L & 35.72 & 59.78 & 72.99 & 80.95 & 85.57 & 30.96 & 51.78 & 65.59 & 76.29 & 80.78 & 78.48 & 91.70 & 95.66 & 97.47 & 98.11 & 40.88 & 62.02 & 73.44 & 80.78 & 86.23 \\
        DS & 35.56 & 58.37 & 71.67 & 79.27 & 83.89 & 30.42 & 49.25 & 62.84 & 71.99 & 78.40 & 78.04 & 91.64 & 96.36 & 98.30 & 98.77 & 39.05 & 59.41 & 72.15 & 80.02 & 84.64 \\
        \midrule
        \multicolumn{21}{c}{\textbf{Set-level Classification (SC)}} \\
        \midrule
        CST + SF & 39.30 & \textbf{63.82} & \textbf{76.83} & \textbf{83.53} & \textbf{87.47} & \textbf{39.04} & \textbf{60.46} & \textbf{74.93} & \textbf{82.20} & \textbf{87.50} & \textbf{80.59} & \textbf{94.12} & \textbf{97.69} & \textbf{98.87} & \textbf{99.39} & \textbf{49.49} & \textbf{71.55} & \textbf{82.91} & \textbf{89.28} & \textbf{92.75} \\
        \textit{$\Delta$} & \textit{-2.28} & \textit{+3.78} & \textit{+3.06} & \textit{+1.29} & \textit{+0.82} & \textit{+5.51} & \textit{+8.14} & \textit{+8.78} & \textit{+5.55} & \textit{+6.25} & \textit{+1.20} & \textit{+1.23} & \textit{+0.69} & \textit{+0.30} & \textit{+0.34} & \textit{+5.87} & \textit{+7.98} & \textit{+7.75} & \textit{+5.43} & \textit{+4.25} \\
        CNN + SF & \textbf{41.58} & 55.39 & 64.14 & 70.39 & 75.16 & 33.53 & 45.95 & 53.35 & 60.89 & 65.12 & 79.39 & 89.80 & 94.14 & 96.31 & 97.38 & 43.62 & 56.99 & 66.28 & 72.61 & 76.79 \\
        ST-S + SF & 35.58 & 60.04 & 73.77 & 82.24 & 86.65 & 31.24 & 52.32 & 65.69 & 74.87 & 81.25 & 79.22 & 92.89 & 97.00 & 98.44 & 99.00 & 42.19 & 63.57 & 75.16 & 83.85 & 88.50 \\
        ST-L + SF & 35.72 & 59.74 & 73.21 & 81.15 & 85.79 & 30.96 & 51.85 & 66.15 & 76.65 & 81.25 & 78.48 & 91.67 & 95.61 & 97.40 & 97.99 & 40.88 & 61.99 & 73.57 & 81.03 & 86.33 \\
        DS + SF & 35.56 & 58.85 & 72.74 & 80.88 & 85.32 & 30.42 & 49.94 & 64.23 & 73.05 & 81.25 & 78.04 & 91.74 & 96.54 & 98.57 & 99.05 & 39.05 & 60.14 & 72.59 & 81.25 & 85.99 \\
        \cmidrule(lr){1-21} 
        CST + LF & \textbf{41.17} & \textbf{65.65} & \textbf{78.93} & \textbf{85.60} & \textbf{89.68} & \textbf{37.37} & \textbf{59.33} & \textbf{72.79} & \textbf{81.55} & \textbf{87.00} & \textbf{80.92} & \textbf{93.73} & \textbf{97.87} & \textbf{99.31} & \textbf{99.72} & \textbf{47.57} & \textbf{70.15} & \textbf{82.26} & \textbf{88.59} & \textbf{92.35} \\
        \textit{$\Delta$} & \textit{+5.14} & \textit{+5.47} & \textit{+4.72} & \textit{+3.49} & \textit{+2.11} & \textit{+6.30} & \textit{+7.60} & \textit{+7.33} & \textit{+6.38} & \textit{+4.86} & \textit{+2.22} & \textit{+1.28} & \textit{+0.90} & \textit{+0.53} & \textit{+0.45} & \textit{+6.15} & \textit{+6.85} & \textit{+5.89} & \textit{+4.65} & \textit{+3.29} \\
        CNN + LF & 35.00 & 58.03 & 71.61 & 79.51 & 85.12 & 31.07 & 50.88 & 65.36 & 74.91 & 81.25 & 78.21 & 92.00 & 96.42 & 98.48 & 99.11 & 40.76 & 63.30 & 75.07 & 83.12 & 89.06 \\
        ST-S  + LF & 36.03 & 60.18 & 74.21 & 82.11 & 86.92 & 31.07 & 51.73 & 65.46 & 75.17 & 82.14 & 78.70 & 92.45 & 96.97 & 98.78 & 99.27 & 40.70 & 61.84 & 76.37 & 83.94 & 88.62 \\
        ST-L + LF & 35.10 & 59.14 & 73.64 & 81.89 & 87.57 & 30.45 & 49.73 & 63.93 & 73.22 & 79.24 & 78.18 & 92.25 & 96.06 & 97.96 & 99.00 & 41.42 & 62.85 & 75.62 & 81.86 & 86.44 \\
        DS + LF & 35.28 & 58.39 & 72.35 & 80.12 & 85.14 & 29.22 & 49.14 & 62.83 & 72.14 & 79.35 & 78.39 & 91.74 & 95.91 & 98.31 & 99.11 & 40.16 & 61.39 & 74.61 & 82.34 & 88.78 \\
        \bottomrule
    \end{tabular}
    }
    \caption{\textbf{Top-1 Test Accuracy (\%) as a function of Set Size (up to 5 images)}. SF and LF are acronyms for Score Fusion and Late Fusion, respectively. Rows labeled as $\Delta$ report the absolute difference in accuracy between CST models and the best-performing baseline.}
    \label{tab:performance_summary}
\end{table*}

%% file: tables/cst_for_tl.tex
\begin{figure}[h!]
    \vfill
    \begin{minipage}{0.40\textwidth}
        \resizebox{\columnwidth}{!}{
            \begin{tabular}{c|c}
                \toprule
                \textbf{CST-15} & \textbf{VGG-19} \\
                \midrule
                28M & 144M \\
                params. & params.\\
                \midrule
                \multicolumn{2}{c}{\textbf{Input: 224 $\times$ 224 RGB images}} \\
                \midrule
                \multicolumn{2}{c}{Conv2D(64, 3) $\times 2$} \\
                \midrule
                \multicolumn{2}{c}{\textbf{Maxpool}} \\
                \midrule
                \multicolumn{2}{c}{Conv2D(128, 3) $\times 2$} \\
                \midrule
                \multicolumn{2}{c}{\textbf{Maxpool}} \\
                \midrule
                SetConv2D(256, 3) $\times 2$ & \multicolumn{1}{c}{Conv2D(256, 3) $\times 4$} \\
                \midrule
                \multicolumn{2}{c}{\textbf{Maxpool}} \\
                \midrule
                SetConv2D(512, 3) $\times 4$ & \multicolumn{1}{c}{Conv2D(512, 3) $\times 4$} \\
                \midrule
                \multicolumn{2}{c}{\textbf{Maxpool}} \\
                \midrule
                SetConv2D(512, 3) $\times 4$ & \multicolumn{1}{c}{Conv2D(512, 3) $\times 4$} \\
                \midrule
                \multicolumn{2}{c}{\textbf{Maxpool}} \\
                \midrule
                \textbf{GAP} & \textbf{Flatten} \\
                \midrule
                - & FC(4096) \\
                - & FC(4096) \\
                \midrule
                \multicolumn{2}{c}{FC(1000)} \\
                \midrule
                \multicolumn{2}{c}{\textbf{Softmax}} \\
                \bottomrule
            \end{tabular}
        }
        \caption{\textbf{CST-15 and \mbox{VGG-19} architectures}. SetConv2D blocks and Conv2D layers specified with ($n_{filters}$, kernel size). All convolutions are implemented with \texttt{same} padding. In SetConv2D blocks, attention heads have dimension 64. Maxpooling is implemented with pool size 2. SetConv2D blocks and Conv2D layers are followed by ReLU6 activation for CST-15 and ReLU for VGG-19.}
        \label{tab:cst15}
    \end{minipage}%
    \hspace{0.50cm}%
    \begin{minipage}{0.55\textwidth}
        
        \resizebox{\textwidth}{!}{ 
            \begin{tabular}{r|c|ccccc}
                \toprule
                \textbf{Model} & \textbf{\# Params.} & \textbf{1} & \textbf{2} & \textbf{3} & \textbf{4} & \textbf{5} \\
                \midrule
                CST-15 & 28M & \textbf{71.37} & \textbf{88.42} & \textbf{92.71} & \textbf{94.62} & \textbf{95.55} \\
                VGG-19 & 144M & 71.24 & - & - & - & - \\
                \bottomrule
            \end{tabular}
        }
        \caption{\textbf{CST-15 and VGG-19 performance on the ImageNet validation set}. Single-crop Classification Accuracy (\%) as a function of the input set size.}
        \label{tab:ilsvrc}
        
        \vspace{1cm}
        
        \includegraphics[width=\textwidth]{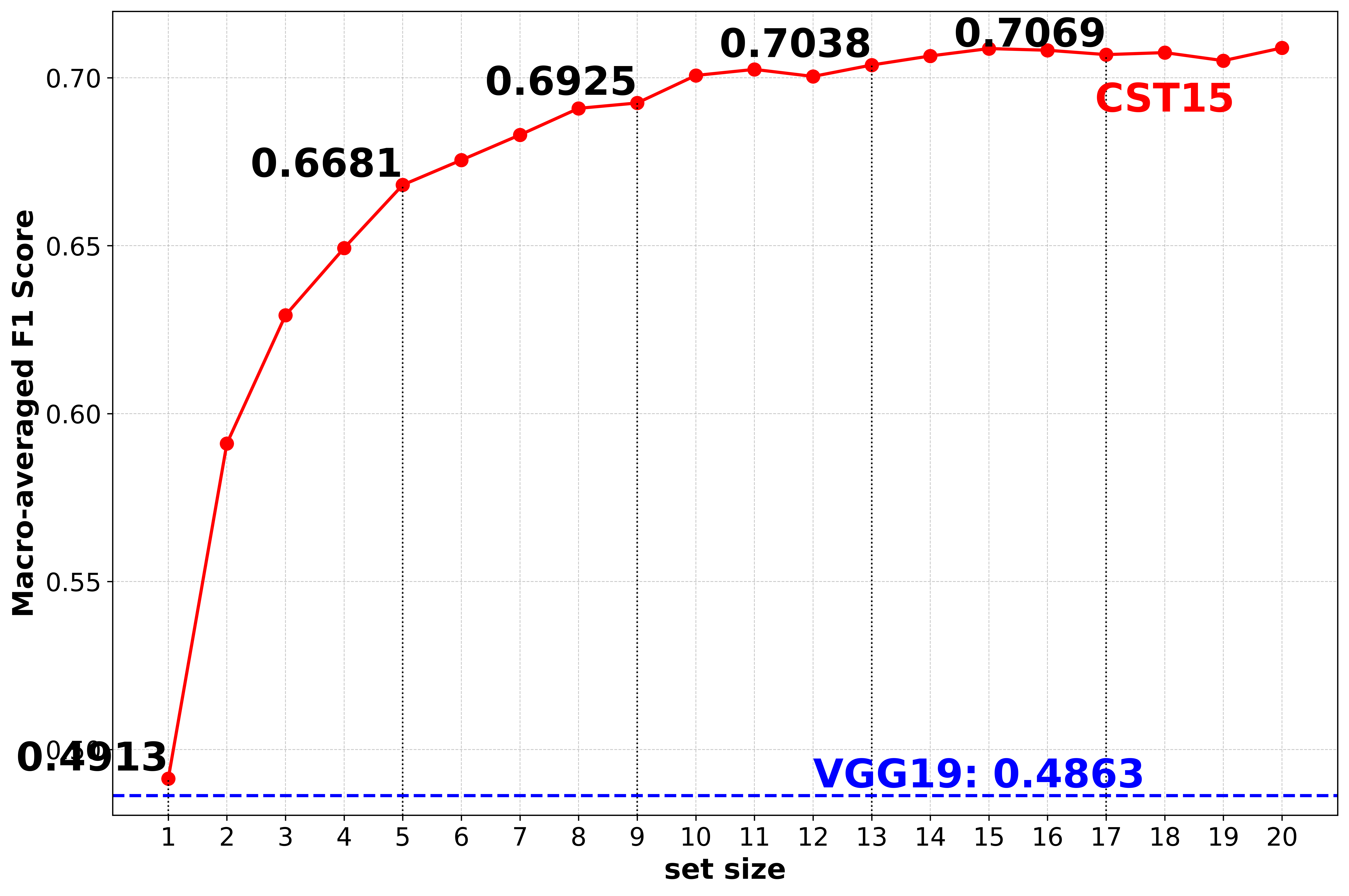}
        \caption{\textbf{Event recognition from personal photos, PEC dataset}. We report the macro-averaged F1 Score as a function of the input set size. Both CST-15 and VGG-19 perform classification at the image level, but while VGG-19 can only processes each image in a set independently of the others, CST-15 leverages the contextual information within the set to improve its image-level classification accuracy.}
        \label{fig:pec_tl}
    \end{minipage}
    \vfill  
\end{figure}

%% file: sec/6_conclusion.tex
\vspace{-0.25cm}
\section{Conclusion}
\label{sec:conclusion}

We introduce the Convolutional Set Transformer (CST), a deep-learning architecture for inference tasks on image sets. Unlike Deep Sets and Set Transformer, CST models the contextual relationships between images in a set directly within convolutional blocks, jointly with feature extraction. Thanks to this, CST delivers superior performance across diverse tasks and datasets and, different from competing architectures, allows for seamless use of explainability tools for CNNs. As a concrete contribution, we present CST-15, the first set-input network pre-trained on the ImageNet dataset, which we release publicly. We further show that CST-15 can be readily adapted to new tasks via Transfer Learning. Looking ahead, future research should explore the many potential applications of CSTs in domains where data is naturally organized as sets of images, including medical imaging, surveillance, e-commerce, and social media.

%% file: appendix.tex
\renewcommand{\thesection}{\Alph{section}}

\vspace*{-1.5cm}
\section{Ablation Study on Combinatorial Training}
\label{sec:ablation_ct}

\setcounter{table}{0}
\renewcommand{\thetable}{\thesection\arabic{table}}
\setcounter{figure}{0}
\renewcommand{\thefigure}{\thesection\arabic{figure}}

Mirroring Table \ref{tab:performance_summary} in the main paper, which reports the performance of CSTs and baseline models trained with Combinatorial Training for the CIC and SC tasks across multiple benchmarks (ImageNet64x64, Tiny ImageNet, CIFAR-10, and CIFAR-100), Table \ref{tab:performance_no_ct} summarizes the results of the same models when trained without Combinatorial Training.\footnote{When CT is not applied, we create sets of three images from the same class \textit{before the training process begins} and use these sets to train the model in every epoch.} 
\input{tables_supplementary/performance_no_ct}
\input{tables_supplementary/performance_diff}

Even when CT is not applied, CSTs outperform the baselines in 55 of 60 test cases. Table \ref{tab:performance_diff} shows, for all models, the relative increase in test accuracy when CT is used as opposed to conventional training. CT is consistently beneficial, with relative improvements often reaching double digits. As a result, CT proves to be an effective training strategy for CSTs and general set-input networks.

\section{Additional Details About the Experiments}
\label{sec:experiment_details}

\setcounter{table}{0}
\renewcommand{\thetable}{\thesection\arabic{table}}
\setcounter{figure}{0}
\renewcommand{\thefigure}{\thesection\arabic{figure}}

In this appendix, we provide additional details about the experiments described in the main paper, including the hyperparameters used for training. \ref{sec:tiny_supp} covers the classification experiments from Section \ref{sec:tiny_experiments} of the paper. \ref{sec:anomaly_detection_supp} focuses on Set Anomaly Detection, as discussed in Section \ref{sec:anomaly}. Finally, \ref{sec:cst15_supp} and \ref{sec:pec_supp} provide further details on CST-15, presented in Section \ref{sec:cst15_pretraining}, and the Transfer Learning experiment from Section \ref{sec:pec_tl}.

\subsection{Classification Experiments}
\label{sec:tiny_supp}

We train both CSTs and all the baseline models---except for standard CNNs, which completely disregard set information---using Combinatorial Training (with \mbox{$n_{min}=2$} and $n_{max}=5$). The batch size is 256 for training on ImageNet64x64, $128$ for Tiny ImageNet, and $64$ for CIFAR-10 and CIFAR-100. We utilize the Adam optimizer \cite{kingma2015adam} (with $\beta_{1}=0.9$ and $\beta_{2}=0.999$) along with a warm-up learning rate schedule. The learning rate starts at $1 \times 10^{-4}$ and linearly increases over the first 5 epochs, reaching $3 \times 10^{-4}$ when training on ImageNet64x64 and $5 \times 10^{-4}$ when training on the other datasets. Notably, CT necessitates low learning rates as concurrent training on sets of different sizes makes the optimization process less stable. For all models, L2 regularization is applied with a coefficient of $5 \times 10^{-4}$. We implement gradient norm clipping for the Late Fusion models, at a threshold of 5. When training CSTs, we impose 10\% dropout probability on attention weights within SetConv2D blocks. For all models, training is terminated when the validation accuracy stops improving for 50 epochs, at which point we select the best-performing model so far.

\begin{table}[h!]
\input{tables_supplementary/anomaly_models.tex}
\end{table}

\subsection{Explainable Set Anomaly Detection}
\label{sec:anomaly_detection_supp}

Table \ref{tab:anomaly_models} presents the architectures tested in the Explainable Set Anomaly Detection experiment. Since it is not possible to have exactly the same number of parameters across different architectures, the CST is compared to a Set Transformer with more parameters (ST-L) and one with fewer parameters (ST-S). The Deep Sets model has fewer parameters than both CST and Set Transformers. We experimented with larger Deep Sets models (i.e., with more Deep Sets equivariant blocks), but they faced substantial convergence issues. We implement Set Attention Blocks and Deep Sets equivariant layers as described in Section \ref{sec:tested_architectures} of the paper.

The models are trained from scratch to detect anomalous images within sets derived from the CelebA dataset. This task is framed as binary classification, where the model predicts for each image whether it is normal or anomalous.
\newpage
The construction of the sets follows the protocol detailed below:

\begin{enumerate}
    \item randomly select two attributes from the 40 available in the CelebA annotations;
    \item randomly select $\lfloor N \times p_{anomaly} \rfloor$ anomalous images that lack both attributes, where $N$ is the set size and $p_{anomaly}$ is the anomaly prevalence rate;
    \item randomly select $N - \lfloor N \times p_{anomaly} \rfloor$ normal images that exhibit both attributes.
\end{enumerate}

We use the default training, validation, and test splits of CelebA. The experiment is conducted at the original CelebA resolution of $178 \times 218$ pixels, without any cropping. During training, sets are continuously generated based on the protocol above, with a set size of 10 and an anomaly prevalence rate that varies randomly between 0 and 0.4. We optimize with Adam ($\beta_{1}=0.9$ and $\beta_{2}=0.999$). The learning rate is set to $1 \times 10^{-4}$. We add L2 regularization with a coefficient of $5 \times 10^{-4}$. For all models, training is terminated when the validation AUPRC stops improving for 10 epochs, at which point we select the best-performing model so far.

\subsection{CST-15 trained on ImageNet}
\label{sec:cst15_supp}

We train CST-15 from scratch using Combinatorial Training at a resolution of $224 \times 224$. Input pre-processing involves scaling pixel values to the $[0,1]$ range, and normalizing each channel with respect to the ImageNet dataset. For scale augmentation, we follow the multi-scale training approach outlined in \cite{simonyan15very}. We optimize with Adam \cite{kingma2015adam} (with $\beta_{1}=0.9$ and $\beta_{2}=0.999$). The learning rate is initially set to $1 \times 10^{-4}$ and reduced as needed in later training stages. We apply L2 regularization with a coefficient of 0.1. The CT hyperparameters, $n_{min}$ and $n_{max}$, are set to 1 and 2, respectively. The batch size is set to 320. The model is trained on a single A100 GPU for 250 epochs.

\subsection{Transfer learning on the PEC dataset}
\label{sec:pec_supp}

For each event class, we reserve six personal albums for validation. The official PEC test split is used for testing. Images are center-cropped and resized at $224 \times 224$. We adapt both CST-15 and VGG-19 by replacing their original 1000-way ImageNet classifier with a new layer for the 14 PEC classes. The backbone networks are kept frozen and only the new classification layer is trained. Following the Set-free Transfer Learning scheme, during the training of the CST-15 model, we input individual images without any set, just as we do for VGG-19. The learning rate is set to $1 \times 10^{-3}$, the L2 regularization coefficient to $5 \times 10^{-4}$, and the batch size to 64. We use Adam \cite{kingma2015adam} for optimization (with $\beta_{1}=0.9$ and $\beta_{2}=0.999$). Training is terminated when the validation accuracy stops improving for 10 epochs, at which point we select the best-performing model so far.

\section{Explainability of CSTs}
\label{sec:explain_cst}

\setcounter{table}{0}
\renewcommand{\thetable}{\thesection\arabic{table}}
\setcounter{figure}{0}
\renewcommand{\thefigure}{\thesection\arabic{figure}}

\begin{figure*}[ht]
    \centering
    \includegraphics[width=0.81\textwidth]{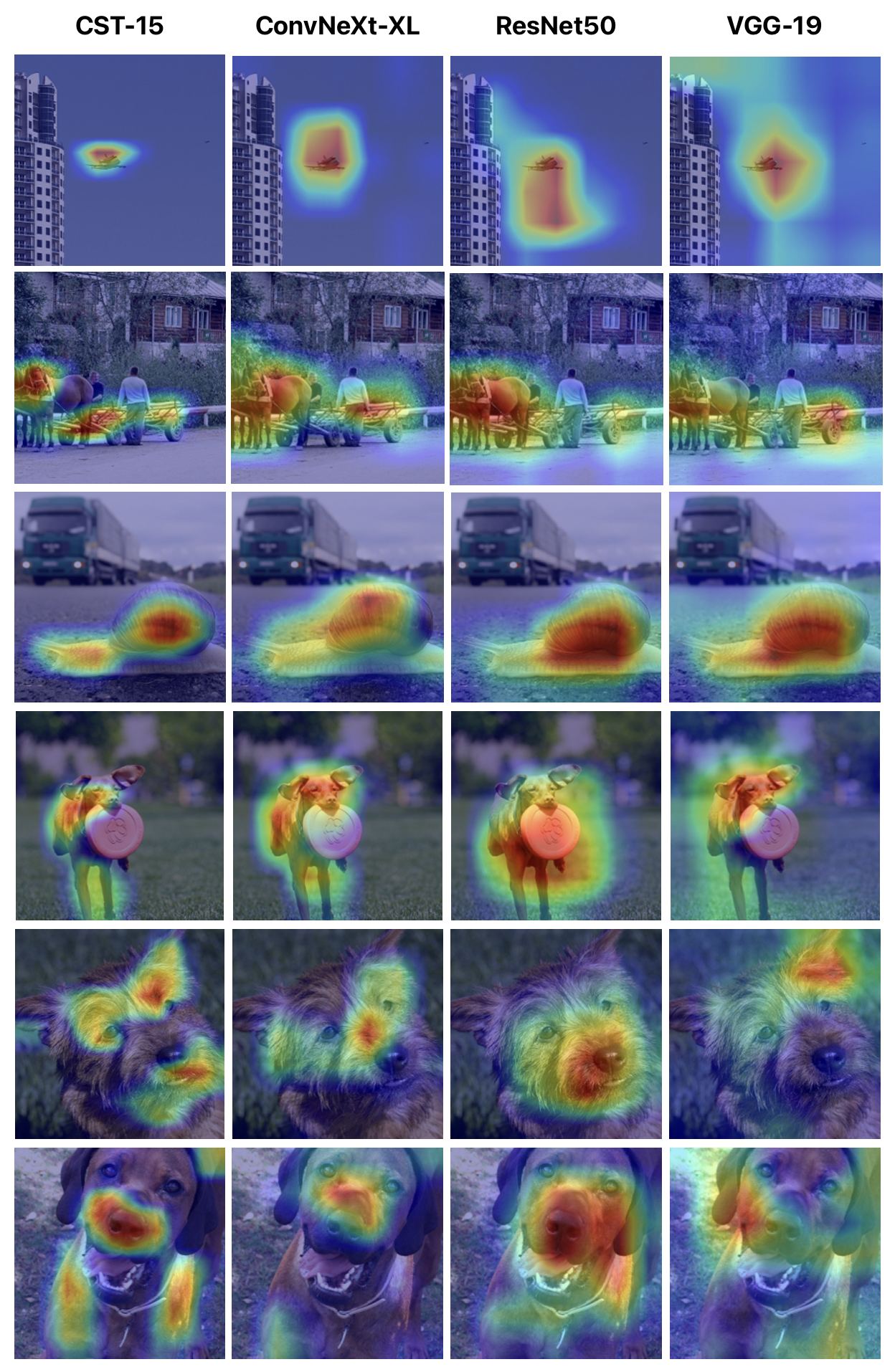}
    \caption{CST-15 Grad-CAMs as opposed to ConvNeXt-XL, ResNet50, and VGG-19 Grad-CAMs. For CST-15, we compute Grad-CAMs at the penultimate SetConv2D block. For the other models, we compute Grad-CAMs at the last layer before GAP or Flattening.}
    \label{fig:grad_cam_comparision}
\end{figure*}

CSTs are readily compatible with standard explainability tools designed for CNNs. Grad-CAMs \cite{Selvaraju_2017_ICCV} computed for CST-15 demonstrate remarkable accuracy, especially when considering the simplicity of the CST-15 architecture. Figure \ref{fig:grad_cam_comparision} presents a qualitative comparison of Grad-CAMs generated for CST-15 (28M params), ConvNeXt-XL (350M params) \cite{liu2022convnet}, ResNet50 (26M params) \cite{he2016deep, he2016identity}, and VGG-19 (144M params) \cite{simonyan15very}. To ensure a fair comparison, we feed isolated images without context to CST-15. Explanations are computed with respect to the ground truth class. \mbox{CST-15} Grad-CAMs are more precise and focused compared to ResNet50 and VGG-19, and comparable or even better than ConvNeXt-XL explanations. 

Consider, for instance, the first image in Figure \ref{fig:grad_cam_comparision}. It depicts a space shuttle being transported by a shuttle carrier aircraft. Even to the human eye, it is difficult to distinguish the shuttle from the carrier aircraft. However, the CST-15 explanation map accurately identifies the space shuttle, distinguishing it from the aircraft. In contrast, the Grad-CAMs generated for the other models are significantly less precise, highlighting a coarse region that encompasses both the shuttle and the carrier aircraft.

\begin{figure*}[ht]
    \centering
    \includegraphics[width=0.50\textwidth]{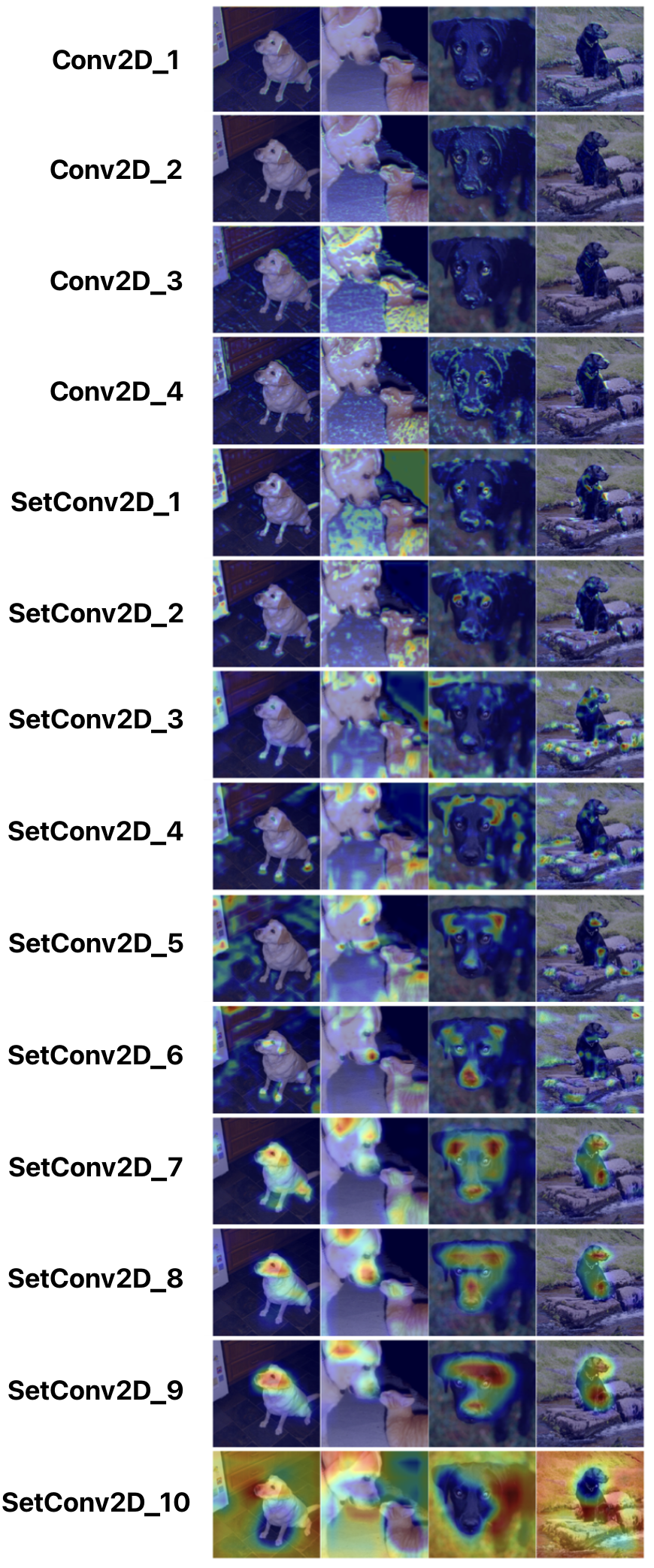}
    \caption{Grad-CAMs for an image set provided as input to CST-15, computed with respect to the ground truth class at various layers of the CST-15 architecture.}
    \label{fig:layer_grad_cams}
\end{figure*}

Importantly, when dealing with CSTs trained for CIC, Grad-CAMs should be computed at the \textit{penultimate SetConv2D} block. As Figure \ref{fig:layer_grad_cams} shows, explanations computed at the final SetConv2D block are negatively affected by the overwhelming importance of contextual information. Recall that CSTs evenly spread contextual information across the spatial dimensions. 

\begin{figure*}[ht]
    \centering
    \includegraphics[width=0.95\textwidth]{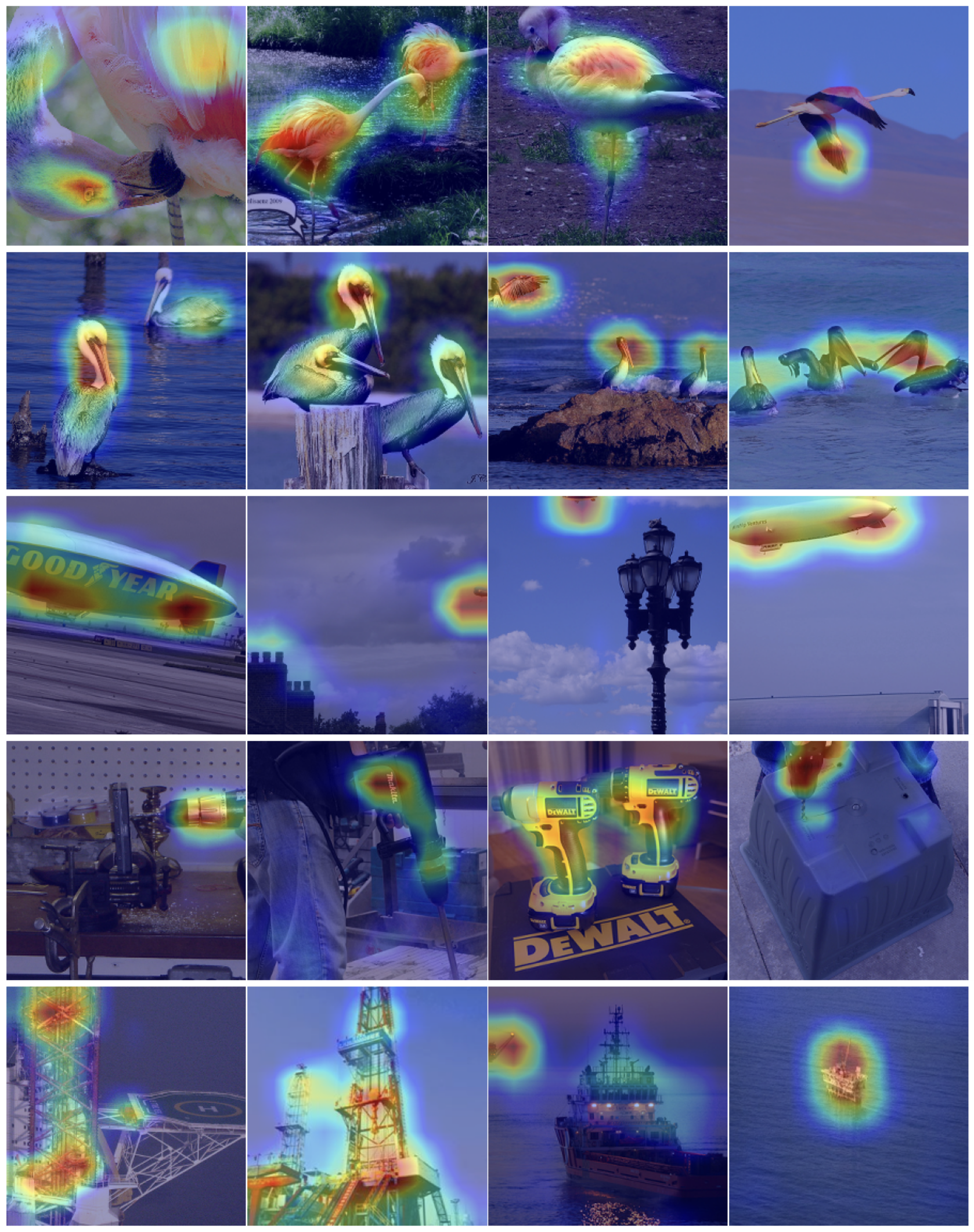}
    \caption{Grad-CAMs for image sets provided as input to CST-15 with respect to the ground truth class. Each row corresponds to a set of images. Grad-CAMs are computed at the penultimate SetConv2D block.}
    \label{fig:grad_cams1}
\end{figure*}

\begin{figure*}[ht]
    \centering
    \includegraphics[width=0.95\textwidth]{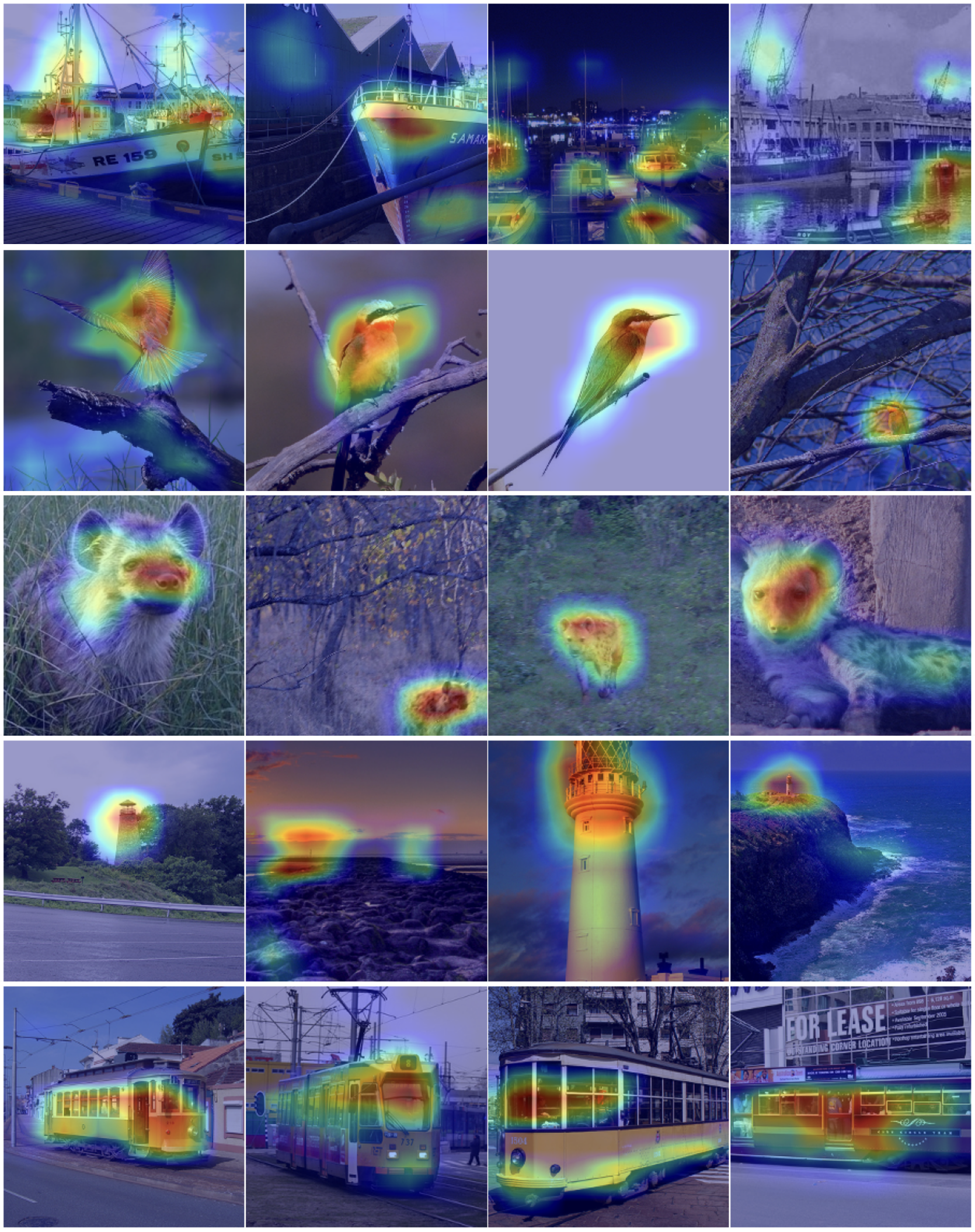}
    \caption{Grad-CAMs for image sets provided as input to CST-15 with respect to the ground truth class. Each row corresponds to a set of images. Grad-CAMs are computed at the penultimate SetConv2D block.}
    \label{fig:grad_cams2}
\end{figure*}

\begin{figure*}[ht]
    \centering
    \includegraphics[width=0.95\textwidth]{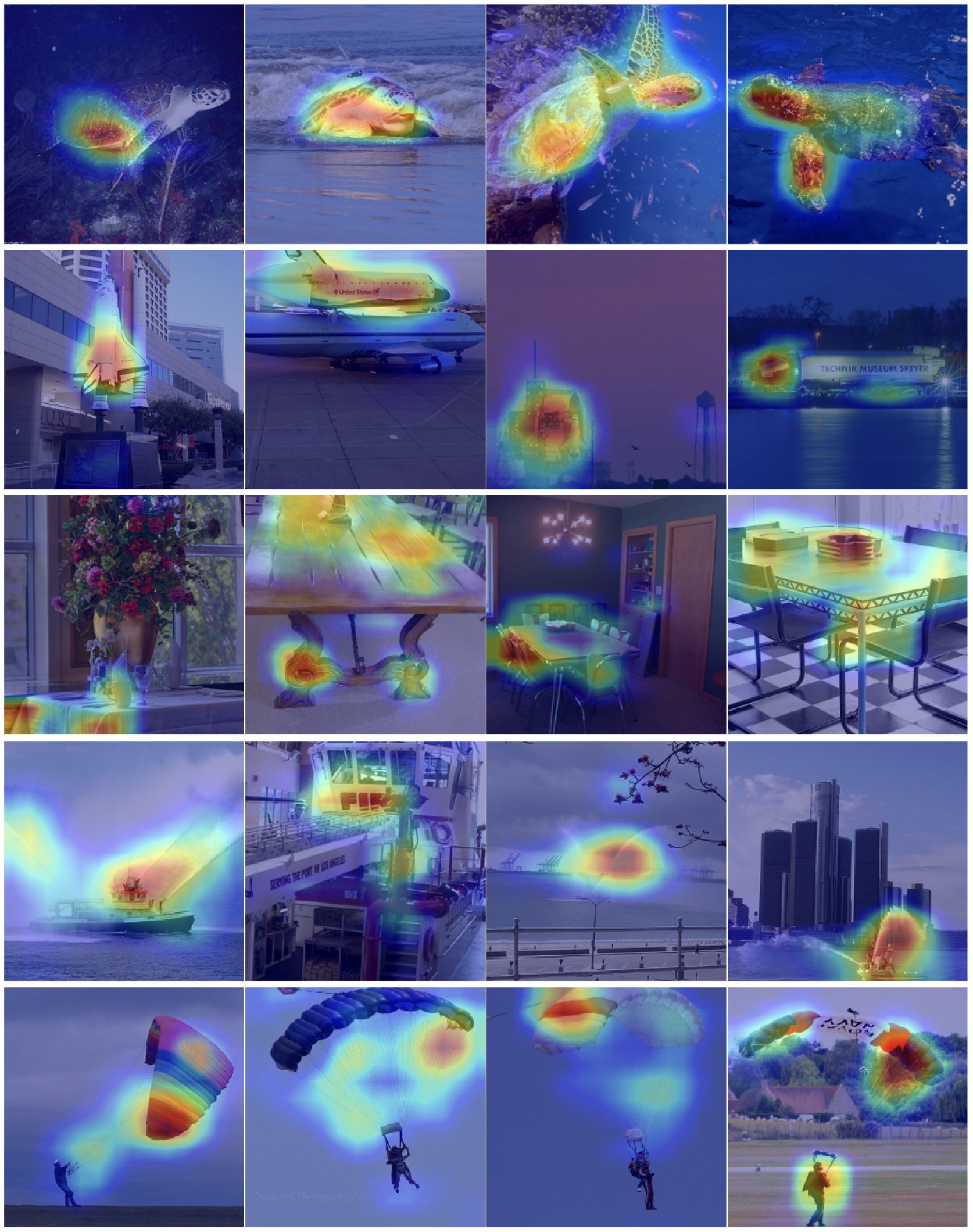}
   \caption{Grad-CAMs for image sets provided as input to CST-15 with respect to the ground truth class. Each row corresponds to a set of images. Grad-CAMs are computed at the penultimate SetConv2D block.}
   \label{fig:grad_cams3}
\end{figure*}

Figures \ref{fig:grad_cams1}, \ref{fig:grad_cams2}, and \ref{fig:grad_cams3} present further Grad-CAMs computed for sets of 4 images fed to CST-15 (each row corresponds to an image set).

%% file: tables_supplementary/performance_no_ct.tex
\begin{table*}[ht]
    \centering
    \huge
    \resizebox{1\textwidth}{!}{ 
    \begin{tabular}{r | ccccc | ccccc | ccccc | ccccc}
        \toprule
        \textbf{Model} & \multicolumn{5}{c}{\textbf{ImageNet64x64}} & \multicolumn{5}{c}{\textbf{Tiny ImageNet}} & \multicolumn{5}{c}{\textbf{CIFAR 10}} & \multicolumn{5}{c}{\textbf{CIFAR 100}} \\
        \cmidrule(lr){2-6} \cmidrule(lr){7-11} \cmidrule(lr){12-16} \cmidrule(lr){17-21}
        & 1 & 2 & 3 & 4 & 5 & 1 & 2 & 3 & 4 & 5 & 1 & 2 & 3 & 4 & 5 & 1 & 2 & 3 & 4 & 5 \\
        \midrule
        \multicolumn{21}{c}{\textbf{Contextualized Image Classification (CIC)}} \\
        \midrule
        CST & \textbf{29.72} & \textbf{52.83} & \textbf{67.54} & \textbf{76.03} & \textbf{80.97} & \textbf{29.56} & \textbf{49.61} & \textbf{61.94} & 71.40 & 75.88 & \textbf{75.73} & \textbf{90.84} & \textbf{96.28} & \textbf{98.07} & \textbf{98.95} & \textbf{40.55} & \textbf{63.43} & \textbf{74.80} & \textbf{82.82} & \textbf{89.10} \\
        $\Delta$  & \textit{+0.97} & \textit{+1.62} & \textit{+2.24} & \textit{+1.83} & \textit{+1.10} & \textit{+0.64} & \textit{+2.17} & \textit{+0.06} & \textit{-0.07} & \textit{-1.81} & \textit{+4.08} & \textit{+3.05} & \textit{+2.20} & \textit{+1.69} & \textit{+0.95} & \textit{+3.52} & \textit{+5.73} & \textit{+4.21} & \textit{+2.74} & \textit{+4.25} \\
        ST-S & 27.85 & 49.77 & 64.59 & 73.61 & 79.53 & 27.98 & 47.44 & 61.88 & \textbf{71.47} & \textbf{77.69} & 71.65 & 87.30 & 93.69 & 96.22 & 97.83 & 37.03 & 57.70 & 70.59 & 80.08 & 84.85 \\
        ST-L & 27.22 & 49.28 & 63.84 & 73.66 & 79.40 & 28.92 & 47.11 & 60.36 & 70.33 & 76.79 & 71.28 & 87.79 & 94.08 & 96.33 & 98.00 & 34.38 & 56.14 & 68.65 & 77.35 & 83.67 \\
        DS & 28.75 & 51.21 & 65.30 & 74.20 & 79.87 & 25.19 & 41.84 & 54.35 & 64.36 & 70.23 & 71.29 & 87.28 & 93.75 & 96.38 & 97.88 & 34.65 & 55.57 & 68.49 & 77.46 & 83.57 \\
        \midrule
        \multicolumn{21}{c}{\textbf{Set-level Classification (SC)}} \\
        \midrule
        CST + SF & \textbf{29.72} & \textbf{52.89} & \textbf{67.60} & \textbf{76.19} & 81.14 & \textbf{29.56} & \textbf{49.65} & 62.11 & \textbf{72.92} & 76.17 & \textbf{75.73}& \textbf{91.00} & \textbf{96.36} & \textbf{98.13} & \textbf{99.00} & \textbf{40.55} & \textbf{63.71} & \textbf{74.61} & \textbf{83.25} & \textbf{88.84} \\
        $\Delta$  & \textit{+0.97} & \textit{+0.95} & \textit{+0.92} & \textit{+0.52} & \textit{-0.34} & \textit{+0.64} & \textit{+2.06} & \textit{-0.42} & \textit{+1.22} & \textit{-2.12} & \textit{+4.08} & \textit{+3.36} & \textit{+1.95} & \textit{+1.25} & \textit{+0.73} & \textit{+3.52} & \textit{+5.40} & \textit{+3.74} & \textit{+2.61} & \textit{+3.24} \\
        ST-S + SF & 27.85 & 49.95 & 64.74 & 74.14 & 79.75 & 27.98 & 47.59 & \textbf{62.53} & 71.70 & \textbf{78.29} & 71.65 & 87.27 & 93.66 & 96.14 & 97.77 & 37.03 & 58.31 & 70.87 & 80.64 & 85.60 \\
        ST-L + SF & 27.22 & 49.26 & 64.03 & 73.85 & 79.74 & 28.92 & 47.18 & 60.45 & 70.62 & 77.51 & 71.28 & 87.64 & 94.17 & 96.22 & 98.27 & 34.38 & 56.29 & 69.17 & 77.65 & 84.15 \\
        DS + SF & 28.75 & 51.94 & 66.68 & 75.67 & \textbf{81.48} & 25.19 & 42.50 & 55.40 & 66.45 & 72.38 & 71.29 & 87.48 & 94.41 & 96.88 & 98.16 & 34.65 & 56.11 & 69.60 & 78.99 & 85.38 \\
        \cmidrule(lr){1-21} 
        CST + LF & \textbf{32.44} & \textbf{56.33} & \textbf{70.90} & \textbf{79.91} & \textbf{84.81} & \textbf{33.15} & \textbf{54.81} & \textbf{69.04} & \textbf{77.52} & \textbf{84.60} & \textbf{75.43} & \textbf{90.75} & \textbf{96.15} & \textbf{98.48} & \textbf{99.11} & \textbf{41.71} & \textbf{64.93} & \textbf{76.99} & \textbf{84.51} & \textbf{90.96} \\
        $\Delta$  & \textit{+3.47} & \textit{+4.05} & \textit{+4.30} & \textit{+3.71} & \textit{+2.86} & \textit{+5.59} & \textit{+7.96} & \textit{+7.45} & \textit{+5.34} & \textit{+6.20} & \textit{+3.71} & \textit{+2.69} & \textit{+1.98} & \textit{+1.65} & \textit{+0.67} & \textit{+5.39} & \textit{+7.04} & \textit{+5.67} & \textit{+4.74} & \textit{+5.30} \\
        CNN + LF & 28.10 & 49.87 & 65.10 & 74.20 & 80.73 & 27.56 & 46.85 & 61.59 & 72.18 & 78.40 & 71.72 & 88.06 & 94.17 & 96.83 & 98.44 & 36.32 & 57.89 & 71.32 & 79.77 & 85.66 \\
        ST-S  + LF & 28.97 & 52.28 & 66.46 & 75.56 & 81.75 & 27.54 & 45.33 & 59.70 & 69.57 & 77.29 & 71.50 & 87.60 & 93.72 & 96.53 & 98.05 & 34.50 & 55.45 & 67.84 & 78.30 & 83.20 \\
        ST-L + LF & 28.53 & 51.39 & 66.60 & 76.20 & 81.95 & 26.62 & 44.88 & 59.02 & 68.32 & 74.11 & 71.47 & 87.79 & 93.96 & 96.22 & 98.10 & 35.10 & 57.11 & 69.63 & 77.69 & 83.82 \\
        DS + LF & 27.86 & 50.06 & 65.08 & 74.30 & 80.89 & 26.23 & 45.58 & 59.34 & 69.49 & 77.68 & 71.06 & 87.66 & 93.90 & 96.57 & 97.88 & 32.84 & 54.61 & 66.99 & 76.52 & 83.48 \\
        \bottomrule
    \end{tabular}
    }
    \caption{\textbf{Top-1 Test Accuracy (\%) as a function of Set Size (up to 5 images). All the models are trained without CT}. SF and LF are acronyms for Score Fusion and Late Fusion, respectively. Rows labeled as $\Delta$ report the absolute difference in accuracy between CST models and the best-performing baseline.}
    \label{tab:performance_no_ct}
\end{table*}

%% file: tables_supplementary/performance_diff.tex
\begin{table*}[ht]
    \centering
    \huge
    \resizebox{1\textwidth}{!}{ 
    \begin{tabular}{r | ccccc | ccccc | ccccc | ccccc}
        \toprule
        \textbf{Model} & \multicolumn{5}{c}{\textbf{ImageNet64x64}} & \multicolumn{5}{c}{\textbf{Tiny ImageNet}} & \multicolumn{5}{c}{\textbf{CIFAR 10}} & \multicolumn{5}{c}{\textbf{CIFAR 100}} \\
        \cmidrule(lr){2-6} \cmidrule(lr){7-11} \cmidrule(lr){12-16} \cmidrule(lr){17-21}
        & 1 & 2 & 3 & 4 & 5 & 1 & 2 & 3 & 4 & 5 & 1 & 2 & 3 & 4 & 5 & 1 & 2 & 3 & 4 & 5 \\
        \midrule
        \multicolumn{21}{c}{\textbf{Contextualized Image Classification (CIC)}} \\
        \midrule
        CST & +32.23 & +20.59 & +13.21 & +9.35 & +7.47 & +32.07 & +20.68 & +19.73 & +13.54 & +14.85 & +6.42 & +3.54 & +1.25 & +0.84 & +0.21 & +22.05 & +11.63 & +10.17 & +6.93 & +3.63 \\
        ST-S & +27.76 & +20.15 & +13.84 & +10.61 & +8.35 & +11.65 & +9.21 & +5.17 & +3.41 & +2.46 & +10.57 & +6.38 & +3.56 & +2.20 & +1.12 & +13.93 & +9.93 & +6.84 & +4.07 & +3.71 \\
        ST-L & +31.23 & +21.31 & +14.33 & +9.90 & +7.77 & +7.05 & +9.91 & +8.66 & +8.47 & +5.20 & +10.10 & +4.45 & +1.68 & +1.18 & +0.11 & +18.91 & +10.47 & +6.98 & +4.43 & +3.06 \\
        DS & +23.69 & +13.98 & +9.75 & +6.83 & +5.03 & +20.76 & +17.71 & +15.62 & +11.86 & +11.63 & +9.47 & +5.00 & +2.78 & +1.99 & +0.91 & +12.70 & +6.91 & +5.34 & +3.30 & +1.28 \\
        \midrule
        \multicolumn{21}{c}{\textbf{Set-level Classification (SC)}} \\
        \midrule
        CST + SF & +32.23 & +20.67 & +13.65 & +9.63 & +7.80 & +32.07 & +21.77 & +20.64 & +12.73 & +14.87 & +6.42 & +3.43 & +1.38 & +0.75 & +0.39 & +22.05 & +12.31 & +11.12 & +7.24 & +4.40 \\
        ST-S + SF & +27.76 & +20.20 & +13.95 & +10.93 & +8.65 & +11.65 & +9.94 & +5.05 & +4.42 & +3.78 & +10.57 & +6.44 & +3.57 & +2.39 & +1.26 & +13.93 & +9.02 & +6.05 & +3.98 & +3.39 \\
        ST-L + SF & +31.23 & +21.27 & +14.34 & +9.88 & +7.59 & +7.05 & +9.90 & +9.43 & +8.54 & +4.83 & +10.10 & +4.60 & +1.53 & +1.23 & -0.28 & +18.91 & +10.13 & +6.36 & +4.35 & +2.59 \\
        DS + SF & +23.69 & +13.30 & +9.09 & +6.89 & +4.71 & +20.76 & +17.51 & +15.94 & +9.93 & +12.25 & +9.47 & +4.87 & +2.26 & +1.74 & +0.91 & +12.70 & +7.18 & +4.30 & +2.86 & +0.71 \\
        \cmidrule(lr){1-21} 
        CST + LF & +26.91 & +16.55 & +11.33 & +7.12 & +5.74 & +12.73 & +8.25 & +5.43 & +5.20 & +2.84 & +7.28 & +3.28 & +1.79 & +0.84 & +0.62 & +14.05 & +8.04 & +6.85 & +4.83 & +1.53 \\
        CNN + LF & +24.56 & +16.36 & +10.00 & +7.16 & +5.44 & +12.74 & +8.60 & +6.12 & +3.78 & +3.64 & +9.05 & +4.47 & +2.39 & +1.70 & +0.68 & +12.22 & +9.35 & +5.26 & +4.20 & +3.97 \\
        ST-S  + LF & +24.37 & +15.11 & +11.66 & +8.67 & +6.32 & +12.82 & +14.12 & +9.65 & +8.05 & +6.28 & +10.07 & +5.54 & +3.47 & +2.33 & +1.24 & +17.97 & +11.52 & +12.57 & +7.20 & +6.51 \\
        ST-L + LF & +23.03 & +15.08 & +10.57 & +7.47 & +6.86 & +14.39 & +10.81 & +8.32 & +7.17 & +6.92 & +9.39 & +5.08 & +2.23 & +1.81 & +0.92 & +18.01 & +10.05 & +8.60 & +5.37 & +3.13 \\
        DS + LF & +26.63 & +16.64 & +11.17 & +7.83 & +5.25 & +11.40 & +7.81 & +5.88 & +3.81 & +2.15 & +10.32 & +4.65 & +2.14 & +1.80 & +1.26 & +22.29 & +12.42 & +11.37 & +7.61 & +6.35 \\
        \bottomrule
    \end{tabular}
    }
    \caption{\textbf{Relative increase (\%) in Top-1 Test Accuracy as a function of Set Size (up to 5 images) when CT is used compared to conventional training.}}
    \label{tab:performance_diff}
\end{table*}

%% file: tables_supplementary/anomaly_models.tex
    \centering
    \huge
    \resizebox{0.55\textwidth}{!}{
    \begin{tabular}{c|c|c|c}
        \toprule
        \textbf{CST} & \textbf{ST-S} & \textbf{ST-L} & \textbf{DS} \\
        \midrule
        27.8M & 27.3M & 28.3M & 19.1M \\
        params. & params. & params. & params. \\
        \midrule
        \multicolumn{4}{c}{\textbf{Input: 178 $\times$ 218  RGB images}} \\
        \midrule
        \multicolumn{4}{c}{Conv2D(64, 3)} \\
        \multicolumn{4}{c}{Conv2D(64, 3)} \\
        \midrule
        \multicolumn{4}{c}{\textbf{Maxpool}} \\
        \midrule
        \multicolumn{4}{c}{Conv2D(128, 3)} \\
        \multicolumn{4}{c}{Conv2D(128, 3)} \\
        \midrule
        \multicolumn{4}{c}{\textbf{Maxpool}} \\
        \midrule
        SetConv2D(256, 3) & \multicolumn{3}{c}{Conv2D(256, 3)} \\
        SetConv2D(256, 3) & \multicolumn{3}{c}{Conv2D(256, 3)} \\
        \midrule
        \multicolumn{4}{c}{\textbf{Maxpool}} \\
        \midrule
        SetConv2D(512, 3) & \multicolumn{3}{c}{Conv2D(512, 3)} \\
        SetConv2D(512, 3) & \multicolumn{3}{c}{Conv2D(512, 3)} \\
        SetConv2D(512, 3) & \multicolumn{3}{c}{Conv2D(512, 3)} \\
        SetConv2D(512, 3) & \multicolumn{3}{c}{Conv2D(512, 3)} \\
        \midrule
        \multicolumn{4}{c}{\textbf{Maxpool}} \\
        \midrule
        SetConv2D(512, 3) & \multicolumn{3}{c}{Conv2D(512, 3)} \\
        SetConv2D(512, 3) & \multicolumn{3}{c}{Conv2D(512, 3)} \\
        SetConv2D(512, 3) & \multicolumn{3}{c}{Conv2D(512, 3)} \\
        SetConv2D(512, 3) & \multicolumn{3}{c}{Conv2D(512, 3)} \\
        \midrule
        \multicolumn{4}{c}{\textbf{Maxpool}} \\
        \midrule
        \multicolumn{4}{c}{\textbf{GAP}} \\
        \midrule
        - & SAB(512) $\times 8$ & SAB(512) $\times 9$ & DS(512) $\times 1$ \\
        \midrule
        \multicolumn{4}{c}{FC(1)} \\
        \midrule
        \multicolumn{4}{c}{\textbf{Sigmoid}} \\
        \bottomrule
    \end{tabular}}
    \caption{\textbf{Architectures tested in the Explainable Set Anomaly Detection experiment.}  SetConv2D blocks and Conv2D layers specified with ($n_{filters}$, kernel size), All convolutions are implemented with \texttt{same} padding. In SetConv2D blocks, attention heads have dimension $64$. Maxpool operations are implemented with pool size $2$. For SAB and Deep Sets blocks, we report the dimension of the output space in parenthesis. In all architectures, we use ReLU activations.}
    \label{tab:anomaly_models}

%% file: main.bbl
\begin{thebibliography}{10}
\expandafter\ifx\csname url\endcsname\relax
  \def\url#1{\texttt{#1}}\fi
\expandafter\ifx\csname urlprefix\endcsname\relax\def\urlprefix{URL }\fi
\expandafter\ifx\csname href\endcsname\relax
  \def\href#1#2{#2} \def\path#1{#1}\fi

\bibitem{collenne2024reset}
J.~Collenne, R.~Iguernaissi, S.~Dubuisson, D.~Merad, Reset: A residual set-transformer approach to tackle the ugly-duckling sign in melanoma detection, in: ICIP, IEEE, 2024, pp. 3186--3191.

\bibitem{black2024multi}
S.~Black, R.~Souvenir, Multi-view classification using hybrid fusion and mutual distillation, in: Proceedings of the IEEE/CVF Winter Conference on Applications of Computer Vision, 2024, pp. 270--280.

\bibitem{khajwal2023post}
A.~B. Khajwal, C.-S. Cheng, A.~Noshadravan, Post-disaster damage classification based on deep multi-view image fusion, Computer-Aided Civil and Infrastructure Engineering 38~(4) (2023) 528--544.

\bibitem{zaheer2017deep}
M.~Zaheer, S.~Kottur, S.~Ravanbakhsh, B.~Poczos, R.~R. Salakhutdinov, A.~J. Smola, Deep sets, NeurIPS 30 (2017).

\bibitem{lee2019set}
J.~Lee, Y.~Lee, J.~Kim, A.~Kosiorek, S.~Choi, Y.~W. Teh, Set transformer: A framework for attention-based permutation-invariant neural networks, in: ICML, PMLR, 2019, pp. 3744--3753.

\bibitem{Selvaraju_2017_ICCV}
R.~R. Selvaraju, M.~Cogswell, A.~Das, R.~Vedantam, D.~Parikh, D.~Batra, Grad-cam: Visual explanations from deep networks via gradient-based localization, in: ICCV, 2017.

\bibitem{liu2015faceattributes}
Z.~Liu, P.~Luo, X.~Wang, X.~Tang, Deep learning face attributes in the wild, in: ICCV, 2015.

\bibitem{vaswani2017attention}
A.~Vaswani, N.~Shazeer, N.~Parmar, J.~Uszkoreit, L.~Jones, A.~N. Gomez, {\L}.~Kaiser, I.~Polosukhin, Attention is all you need, NeurIPS (2017).

\bibitem{russakovsky2015imagenet}
O.~Russakovsky, J.~Deng, H.~Su, J.~Krause, S.~Satheesh, S.~Ma, Z.~Huang, A.~Karpathy, A.~Khosla, M.~Bernstein, et~al., Imagenet large scale visual recognition challenge, International journal of computer vision 115 (2015) 211--252.

\bibitem{bossard13}
L.~Bossard, M.~Guillaumin, L.~Van~Gool, Event recognition in photo collections with a stopwatch hmm, in: ICCV, 2013.

\bibitem{chenstacking}
Z.~Chen, X.~Zhu, D.~Su, J.~C. Chuang, Stacking deep set networks and pooling by quantiles, in: ICML, 2024.

\bibitem{wagstaff2019limitations}
E.~Wagstaff, F.~Fuchs, M.~Engelcke, I.~Posner, M.~A. Osborne, On the limitations of representing functions on sets, in: ICML, PMLR, 2019, pp. 6487--6494.

\bibitem{bueno2021representation}
C.~Bueno, A.~Hylton, On the representation power of set pooling networks, NeurIPS 34 (2021) 17170--17182.

\bibitem{wagstaff2022universal}
E.~Wagstaff, F.~B. Fuchs, M.~Engelcke, M.~A. Osborne, I.~Posner, Universal approximation of functions on sets, Journal of Machine Learning Research 23~(151) (2022) 1--56.

\bibitem{naderializadeh2021pooling}
N.~Naderializadeh, J.~F. Comer, R.~Andrews, H.~Hoffmann, S.~Kolouri, Pooling by sliced-wasserstein embedding, NeurIPS 34 (2021) 3389--3400.

\bibitem{bartunov2022equilibrium}
S.~Bartunov, F.~B. Fuchs, T.~P. Lillicrap, Equilibrium aggregation: Encoding sets via optimization, in: Uncertainty in Artificial Intelligence, PMLR, 2022, pp. 139--149.

\bibitem{zhang2019fspool}
Y.~Zhang, J.~Hare, A.~Pr{\"u}gel-Bennett, Fspool: Learning set representations with featurewise sort pooling, arXiv preprint arXiv:1906.02795 (2019).

\bibitem{qi2017pointnet}
C.~R. Qi, H.~Su, K.~Mo, L.~J. Guibas, Pointnet: Deep learning on point sets for 3d classification and segmentation, in: CVPR, 2017, pp. 652--660.

\bibitem{qi2017pointnet++}
C.~R. Qi, L.~Yi, H.~Su, L.~J. Guibas, Pointnet++: Deep hierarchical feature learning on point sets in a metric space, NeurIPS 30 (2017).

\bibitem{maron2020learning}
H.~Maron, O.~Litany, G.~Chechik, E.~Fetaya, On learning sets of symmetric elements, in: International conference on machine learning, PMLR, 2020, pp. 6734--6744.

\bibitem{su2015multi}
H.~Su, S.~Maji, E.~Kalogerakis, E.~Learned-Miller, Multi-view convolutional neural networks for 3d shape recognition, in: ICCV, 2015, pp. 945--953.

\bibitem{wei2020view}
X.~Wei, R.~Yu, J.~Sun, View-gcn: View-based graph convolutional network for 3d shape analysis, in: CVPR, 2020, pp. 1850--1859.

\bibitem{liang2021mhfp}
Q.~Liang, Q.~Li, L.~Zhang, H.~Mi, W.~Nie, X.~Li, Mhfp: Multi-view based hierarchical fusion pooling method for 3d shape recognition, Pattern Recognition Letters 150 (2021) 214--220.

\bibitem{liu2022vfmvac}
Z.~Liu, Y.~Zhang, J.~Gao, S.~Wang, Vfmvac: View-filtering-based multi-view aggregating convolution for 3d shape recognition and retrieval, Pattern Recognition 129 (2022) 108774.

\bibitem{yang2025webly}
X.~Yang, Q.~Guo, W.~Chen, M.~Song, Webly supervised 3d shape recognition, Pattern Recognition 158 (2025) 110982.

\bibitem{lin2025multi}
Y.~Lin, X.~Dou, X.~Luo, Z.~Wu, C.~Liu, T.~Luo, J.~Wen, B.~W.-k. Ling, Y.~Xu, W.~Wang, Multi-view diabetic retinopathy grading via cross-view spatial alignment and adaptive vessel reinforcing, Pattern Recognition 164 (2025) 111487.

\bibitem{sun2019multi}
L.~Sun, J.~Wang, Z.~Hu, Y.~Xu, Z.~Cui, Multi-view convolutional neural networks for mammographic image classification, IEEE Access 7 (2019) 126273--126282.

\bibitem{liu2018multi}
X.~Liu, F.~Hou, H.~Qin, A.~Hao, Multi-view multi-scale cnns for lung nodule type classification from ct images, Pattern Recognition 77 (2018) 262--275.

\bibitem{van2021multi}
G.~Van~Tulder, Y.~Tong, E.~Marchiori, Multi-view analysis of unregistered medical images using cross-view transformers, in: Medical Image Computing and Computer Assisted Intervention--MICCAI 2021: 24th International Conference, Strasbourg, France, September 27--October 1, 2021, Proceedings, Part III 24, Springer, 2021, pp. 104--113.

\bibitem{zhang2024attention}
Y.~Zhang, Y.~Cao, T.~Zhang, W.~Shen, Attention fusion reverse distillation for multi-lighting image anomaly detection, arXiv preprint arXiv:2406.04573 (2024).

\bibitem{do2017plant}
T.-B. Do, H.-H. Nguyen, H.~Vu, T.-L. Le, et~al., Plant identification using score-based fusion of multi-organ images, in: 2017 9th International conference on knowledge and systems engineering (KSE), IEEE, 2017, pp. 191--196.

\bibitem{lee2018multi}
S.~H. Lee, C.~S. Chan, P.~Remagnino, Multi-organ plant classification based on convolutional and recurrent neural networks, IEEE TIP 27~(9) (2018) 4287--4301.

\bibitem{seeland2021multi}
M.~Seeland, P.~M{\"a}der, Multi-view classification with convolutional neural networks, Plos one (2021).

\bibitem{deng2009imagenet}
J.~Deng, W.~Dong, R.~Socher, L.-J. Li, K.~Li, L.~Fei-Fei, Imagenet: A large-scale hierarchical image database, in: 2009 IEEE conference on computer vision and pattern recognition, Ieee, 2009, pp. 248--255.

\bibitem{fellbaum1998wordnet}
C.~Fellbaum, WordNet: An electronic lexical database, MIT press, 1998.

\bibitem{chrabaszcz2017downsampled}
P.~Chrabaszcz, I.~Loshchilov, F.~Hutter, A downsampled variant of imagenet as an alternative to the {CIFAR} datasets, arXiv preprint arXiv:1707.08819 (2017).

\bibitem{le2015tiny}
Y.~Le, X.~Yang, Tiny imagenet visual recognition challenge (2015).

\bibitem{krizhevsky2009learning}
A.~Krizhevsky, Learning multiple layers of features from tiny images (2009).

\bibitem{ba2016layer}
J.~L. Ba, Layer normalization, arXiv preprint arXiv:1607.06450 (2016).

\bibitem{simonyan15very}
K.~Simonyan, A.~Zisserman, Very deep convolutional networks for large-scale image recognition, in: ICLR, 2015.

\bibitem{krizhevsky2010convolutional}
A.~Krizhevsky, \href{https://www.cs.toronto.edu/~kriz/conv-cifar10-aug2010.pdf}{Convolutional deep belief networks on {CIFAR-10}} (2010).
\newline\urlprefix\url{https://www.cs.toronto.edu/~kriz/conv-cifar10-aug2010.pdf}

\bibitem{howard2017mobilenets}
A.~G. Howard, Mobilenets: Efficient convolutional neural networks for mobile vision applications, arXiv preprint arXiv:1704.04861 (2017).

\bibitem{sandler2018mobilenetv2}
M.~Sandler, A.~Howard, M.~Zhu, A.~Zhmoginov, L.-C. Chen, Mobilenetv2: Inverted residuals and linear bottlenecks, in: CVPR, 2018, pp. 4510--4520.

\bibitem{kingma2015adam}
D.~P. Kingma, J.~Ba, Adam: A method for stochastic optimization, in: ICLR, 2015.

\bibitem{liu2022convnet}
Z.~Liu, H.~Mao, C.-Y. Wu, C.~Feichtenhofer, T.~Darrell, S.~Xie, A convnet for the 2020s, in: CVPR, 2022, pp. 11976--11986.

\bibitem{he2016deep}
K.~He, X.~Zhang, S.~Ren, J.~Sun, Deep residual learning for image recognition, in: CVPR, 2016, pp. 770--778.

\bibitem{he2016identity}
K.~He, X.~Zhang, S.~Ren, J.~Sun, Identity mappings in deep residual networks, in: ECCV, 2016, pp. 630--645.

\end{thebibliography}
